\documentclass{article}
\usepackage[final]{colm2026_conference}
\usepackage[T1]{fontenc}
\usepackage{microtype}
\usepackage{booktabs}
\usepackage{graphicx}
\usepackage{hyperref}
\usepackage{url}
\usepackage{amsmath}
\usepackage{amssymb}
\usepackage{multirow}
\usepackage{enumitem}
\usepackage{placeins}
\usepackage{caption}
\usepackage{xcolor}
\usepackage{lineno}

\definecolor{darkblue}{rgb}{0,0,0.5}
\hypersetup{
  colorlinks=true,
  citecolor=darkblue,
  linkcolor=darkblue,
  urlcolor=darkblue
}

\title{CoMem: Reusing Transformer Depth across Queries with
Persistent Intermediate Residuals}

\author{%
  \textbf{Hanzuo Liu}$^{\dag\ddag}$ \quad
  \textbf{Xuan Qi}$^{\dag}$ \quad
  \textbf{Chunyu Liu}$^{\dag}$ \quad
  \textbf{Haotian Zhong}$^{\dag}$ \\
  \textbf{Yulong Wang}$^{\ddag}$ \quad
  \textbf{Key}$^{\ddag}$ \quad
  \textbf{Rayying}$^{\ddag}$ \quad
  \textbf{Alex Lamb}$^{\dag}$ \quad
  \textbf{Mingyu Gao}$^{\dag}$ \\
  $^{\dag}$\text{Tsinghua University} \quad
  $^{\ddag}$\text{Tencent} \\
  \texttt{\{lhz24@mails., gaomy@\}tsinghua.edu.cn} \\
  \url{https://github.com/liuhanzuo/Mixture-of-Memory}%
}

\begin{document}
\maketitle
\lhead{Published at the COLM 2026 Workshop on Efficient Reasoning.}

\begin{abstract}
Repeated queries over shared documents repeatedly execute the same lower
transformer layers. We introduce \textbf{CoMem}, which makes split depth $j$ an
explicit reusable-context axis: write one depth-$j$ residual per token, select a
bounded chunk set, and resume only layers $[j{:}L)$. Among document-reuse
systems we are aware of, CoMem jointly makes split depth a tunable serving axis
and isolates it with a matched $j{=}0$ endpoint. On Qwen3-8B, $j{=}12$ reduces
selected-pack Read from 931.9 to 664.4\,ms ($1.403\times$), with a 3.12-point
RULER cost (95\% CI $[2.36,3.93]$); a continuous-prefix oracle recovers the full
gap. The resulting depth axis quantifies a quality--latency--storage trade-off;
a separate same-adapter, Write-inclusive pipeline is $2.74\times$ faster.
Equal-latency raw replay leads by 11.56 points with BM25, directly measuring an
applicability boundary of prepaid depth rather than hiding it. CoMem stores
8\,KiB/token versus 144\,KiB/token for a protocol-aligned same-Qwen3
CacheBlend-style diagnostic; the cohorts and adaptation budgets are not
matched. A context--position
factorization identifies missing lower-layer document context as the dominant
tested multikey error, and a 32-token overlap raises 92.5 to 98.5 without
increasing persistent bytes or per-query Read. CoMem opens transformer depth as
a measurable, tunable dimension for repeated-query long-context serving.
\end{abstract}

\section{Introduction}
\label{sec:intro}

\begin{figure}[t]
\centering
\IfFileExists{figures/fig1_comem_autofigure.pdf}
  {\includegraphics[width=0.96\textwidth]{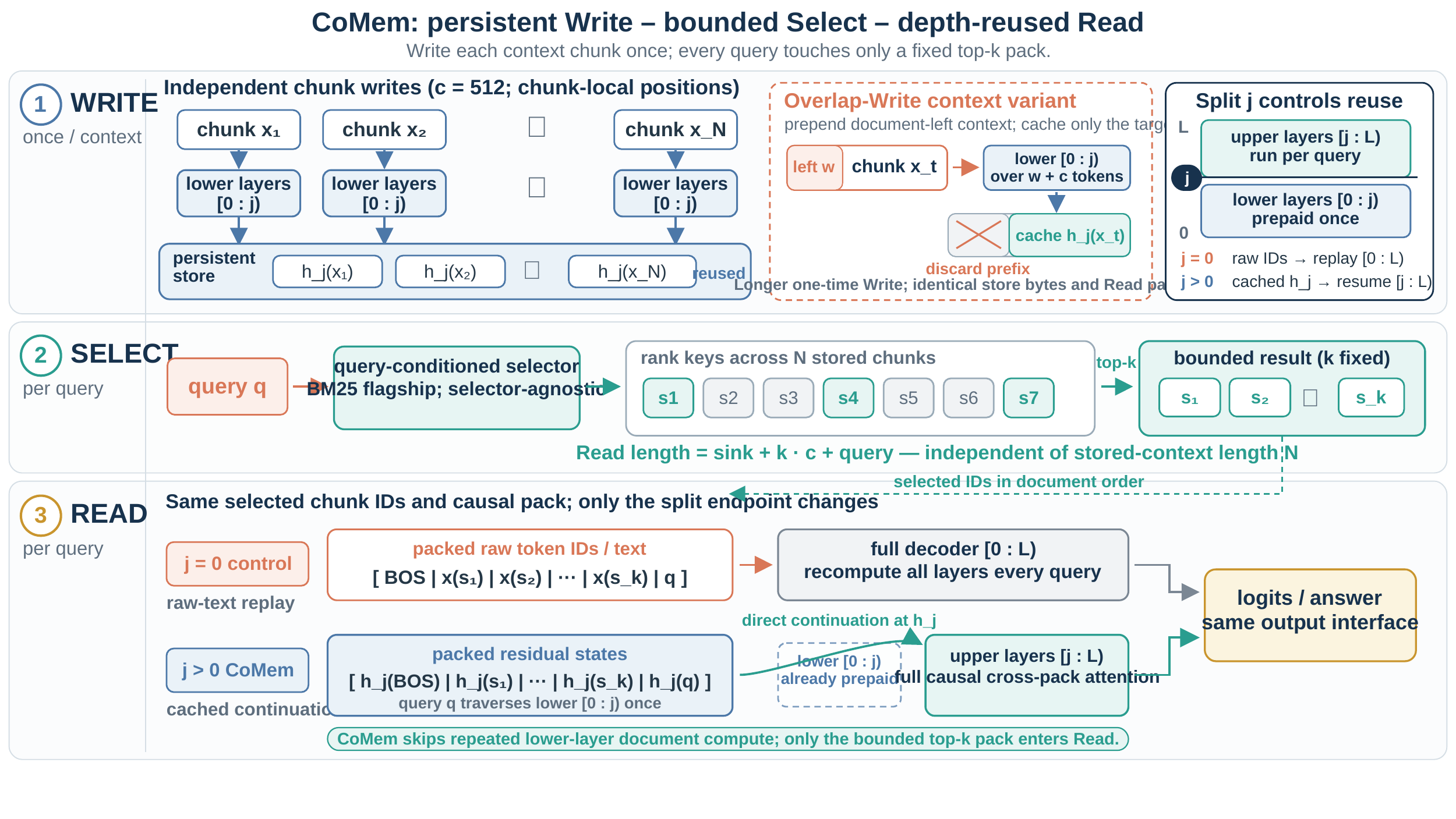}}
  {\includegraphics[width=0.96\textwidth]{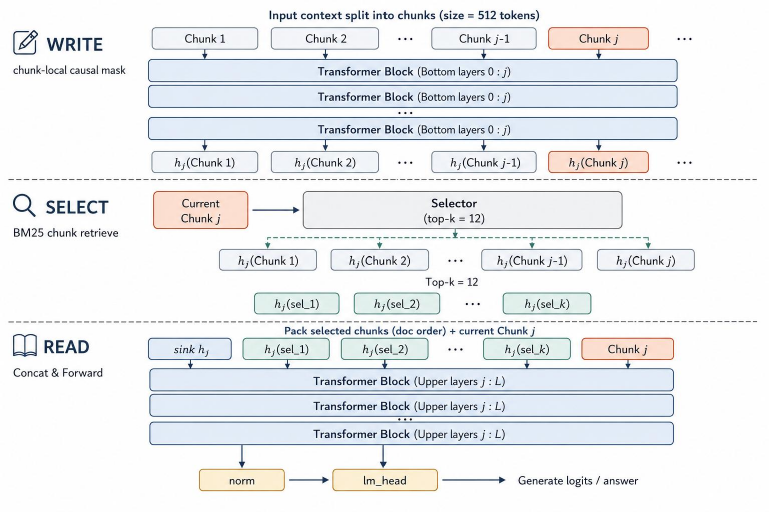}}
\caption{\textbf{CoMem turns transformer depth into a cross-query reuse axis.}
\textsc{Write} prepays layers $[0{:}j)$ once per document chunk,
\textsc{Select} retrieves a bounded working set, and \textsc{Read} resumes
$[j{:}L)$ with the query present. The matched $j{=}0$ endpoint uses the same
selected chunks, order, sink, mask, and adapter while replaying all layers,
thereby isolating the incremental depth effect. Overlap-Write supplies a small
amount of left document context without enlarging the stored object or Read
pack.}
\label{fig:teaser}
\end{figure}

Long-context applications repeatedly ask new questions about the same manuals,
repositories, histories, or document collections. Raw-text retrieval reduces
the number of online tokens, but every selected token still traverses every
transformer layer~\citep{rag,xu2024retrievallong}. Reusable-context systems move
this work offline: CacheBlend, TurboRAG, EPIC, APE, and related systems cache
layer-wise KV and repair context or position dependencies
~\citep{cacheblend,turborag,epic,ape,cachecraft}. Yet these interfaces largely
fix what is stored and how much model depth has already been executed. This
leaves a basic systems question unanswered: \emph{how much transformer depth
should each reusable document object prepay?}

We answer this question with \textbf{CoMem}, which exposes the split $j$ between
offline document processing and online query processing. \textsc{Write} runs
each chunk through layers $[0{:}j)$ and stores its residual $h_j$.
\textsc{Select} retrieves top-$k$ chunks. \textsc{Read} packs the selected
residuals with the query and executes only $[j{:}L)$ with full cross-pack
attention. The $j{=}0$ endpoint stores token IDs and replays the identical pack
through all layers. Varying $j$ therefore trades persistent bytes and
write-once computation against per-query model depth while holding retrieved
evidence fixed.

Prior activation-reuse methods cache a chosen intermediate representation for
repeated downstream runs~\citep{readonce,embeddingrecycling}; LLMCache maintains
per-layer activation banks and semantically matches similar inputs at arbitrary
layers~\citep{llmcache}; HCache stores hidden states across layers to reconstruct
layer-wise KV~\citep{hcache}; and PIC/modular-cache systems store or learn
layer-wise KV objects~\citep{cacheblend,epic,kvpacket,cartridgesbase}. CoMem's
specific conjunction is a persistent document object at one chosen split,
direct decoder-suffix continuation, and an identical-evidence $j{=}0$ endpoint.
This formulation makes prepaid depth directly measurable rather than an
implementation constant.

The matched result establishes a clear quality--latency trade-off. On
Qwen3-8B, moving from $j{=}0$ to $j{=}12$ reduces selected-pack Read from
931.9 to 664.4\,ms, a $1.403\times$ speedup, while the paired 15-cell RULER
macro changes from 99.19 to 96.07 (gap 3.12, 95\% CI $[2.36,3.93]$). A
rank-32 self-distillation adapter makes the split interface effective while
leaving the backbone frozen. When bounded selection and depth reuse are
combined, a separate H20 Pareto harness reduces store-ready 128k online prefill
from 50.59 to 1.32\,s. This operating point compares stock,
LoRA-off dense prefill with adapted CoMem and excludes one-time Write and
external fetch; it is therefore not a same-adapter estimate. A different
same-adapter, Write-inclusive L20A pipeline improves from 6.035 to 2.202\,s
($2.74\times$). The $1.403\times$ result isolates prepaid depth; the other two
measure composed deployment paths under explicitly different boundaries.

CoMem further provides a unified measurement of a
single-residual reusable object across quality, per-query latency, persistent
storage, one-time Write, and empirical break-even. The residual costs
8\,KiB/token---18$\times$ smaller than full bf16 KV for Qwen3-8B, but much
larger than text. In the write-once serving harness, CPU-pinned storage breaks
even after 8.9--10.9 repeated queries at 32k for generations up to 128 tokens;
at 128k and one generated token, GPU/CPU placement breaks even after
25.8--27.6 queries. The full serving grid extends these measurements to 1M
stored tokens and four generation lengths (Table~\ref{tab:serving-crossover}).

The experiments yield two further scientific findings. First, a
continuous-prefix $h_{12}$ oracle exactly recovers the 3.12-point matched gap,
locating the attainable loss at the reusable Write interface rather than the
capacity of the upper decoder. A $2{\times}2$ context--position factorization
then identifies missing lower-layer document context as the dominant tested
factor on paired 8k/16k multikey instances: full document context raises 92.5
to 100.0, while remapping positions alone lowers it to 88.0. A deployable
32-token overlap reaches 98.5 with unchanged persistent bytes and per-query
Read. Second, the equal-latency comparison delineates where depth reuse
applies. Under the common iterative-BM25 ranking, raw replay leads CoMem by
11.56 points: spending the saved latency on full-depth replay is preferable for
this tested mixture. When only replay uses frozen BGE, its 1.00-point edge is
unresolved under hierarchical cell resampling. These statistical boundaries
make retrieval policy a first-order coordinate of the deployment frontier.

\paragraph{Scope.}
Our central claims concern the matched depth axis, measured amortization, and
the tested Write-context mechanism. CoMem targets repeated queries over stable
corpora with open-weight models; the nearest PIC and learned modular-KV systems
use different backbones and measurement boundaries. We therefore pair the
broader conceptual comparison with one same-Qwen3 minimal-faithful
CacheBlend-style experiment rather than transfer unmatched published numbers.
The full set of deployment, generalization, and statistical limits is
consolidated in the dedicated Limitations section.

\paragraph{Contributions.}
\begin{itemize}[leftmargin=1.2em,itemsep=1pt,topsep=2pt]
\item \textbf{A depth-partitioned memory abstraction.} CoMem combines
independently writable intermediate residuals, bounded content selection, and
direct upper-layer replay, with split depth jointly controlling storage,
fidelity, and query-time work.
\item \textbf{A matched $j{=}0$ endpoint design.} Identical-pack replay cleanly
separates the depth-reuse effect from retrieval and packing; oracle and
factorized controls localize the deployable fidelity gap to independent chunk
writing.
\item \textbf{A quantified quality--storage--latency frontier.} We measure
depth splits, selectors, and deployment scenarios, including a qualified
same-backbone prior-art comparison and an equal-latency boundary that defines
when depth reuse is and is not beneficial.
\end{itemize}

\section{Related Work}
\label{sec:related}

We organize prior work by the persistent object and the model computation that
remains after retrieval (Table~\ref{tab:priorart}). This comparison exposes the
specific conjunction studied by CoMem: a persistent document residual at one
chosen split, direct decoder-suffix execution, and an identical-evidence
$j{=}0$ endpoint for measuring the incremental depth effect.

\begin{table}[t]
\centering
\small
\setlength{\tabcolsep}{5pt}
\renewcommand{\arraystretch}{1.05}
\begin{tabular}{@{}p{0.23\textwidth}p{0.22\textwidth}p{0.22\textwidth}p{0.21\textwidth}@{}}
\toprule
\textbf{Family} & \textbf{Stored state} & \textbf{Tunable axis} &
\textbf{Online work} \\
\midrule
Text replay & token IDs & selection budget & all layers \\
Activation reuse & intermediate / multi-layer states & matching / restoration & downstream or rebuilt layers \\
Full-depth KV reuse & cached or learned KV & linking / repair & repair + decode \\
\textbf{CoMem} & one residual / token & \textbf{split depth $j$} & suffix $[j{:}L)$ \\
\bottomrule
\end{tabular}
\caption{\textbf{Compact positioning of reusable-context interfaces.}
CoMem makes prepaid depth tunable and provides a matched $j{=}0$ endpoint;
detailed distinctions appear in Appendix~\ref{app:systems}.}
\label{tab:priorart}
\end{table}

\paragraph{Reusable intermediate representations.}
ReadOnce Transformers construct compressed, task-independent document
representations for repeated downstream use, while Embedding Recycling caches
an intermediate encoder layer and trains adapters above it
~\citep{readonce,embeddingrecycling}. These works establish that intermediate
text representations can be persistent and useful, but their main trade-off is
representation compression or downstream adaptation; the cached layer is a
chosen architecture setting rather than a measured decoder-serving axis.
CoMem operates in an autoregressive decoder, keeps one residual per selected
token, and varies the suffix depth executed for each query.

LLMCache caches intermediate activations in an independent bank for every
transformer layer and uses semantic fingerprints to reuse a matching state for
similar inputs~\citep{llmcache}. It therefore makes cache behavior
layer-dependent and supports arbitrary layers, but it does not define one
document residual split followed by a fixed native suffix, nor an
identical-evidence $j{=}0$ endpoint. HCache and KV-Direct are closer at the state
level. HCache saves hidden states
throughout the model and reconstructs each layer's KV for fast restoration;
KV-Direct reconstructs layer-wise KV from residual checkpoints
~\citep{hcache,kvdirect}. Their goal is efficient restoration of the standard
layer-wise KV state. CoMem chooses one split residual and executes the remaining
transformer blocks directly, which creates a different storage--depth--quality
trade-off and admits the matched $j{=}0$ control.

\paragraph{Full-depth chunk-KV reuse.}
Prompt Cache reuses exact-match prompt modules, while Block-Attention and
TurboRAG independently precompute chunk KV and restore positional consistency
when modules are assembled~\citep{promptcache,blockattention,turborag}.
CacheBlend adds selective recomputation to repair cross-chunk dependencies;
RAGCache and Cache-Craft optimize management and repair around the same
full-depth object~\citep{cacheblend,ragcache,cachecraft}. These systems avoid
most document recomputation by storing KV at every layer. CoMem instead makes
\emph{cached depth} the adjustable resource: one $h_j$ residual replaces the
full-depth KV object, and layers $[j{:}L)$ reconstruct query-conditioned state.
Our same-Qwen3 comparison (Table~\ref{tab:cacheblend-baseline}) makes this
difference empirical: a minimal faithful CacheBlend-style implementation with
global RoPE reindexing and up to 18\% selective recomputation reaches
67.80--74.70 RULER, 48.50--49.83 BABILong qa5, and 16.77--17.37 LoCoMo
substring accuracy, versus CoMem's 97.05, 68.67, and 23.36 while storing
$18\times$ fewer bytes/token. This comparison isolates a full-depth
chunk-KV alternative under one backbone and evidence pack; it does not replace
each system's native serving implementation.

\paragraph{PIC and learned modular document caches.}
EPIC and MEPIC formalize position-independent caching and reduce link-time
recomputation; APE encodes contexts in parallel and realigns attention
~\citep{epic,mepic,ape}. Fusion RAG Cache further studies cache selection and
fusion at serving scale~\citep{fusionrag}. Learned modular objects include KV
Packet's compiled context-independent KV and Cartridges' self-studied reusable
KV representations~\citep{kvpacket,cartridgesbase}. Cartridges at Scale and
SemPIC extend learned per-document KV retrieval~\citep{cartridges,sempic}.
They optimize how a full-depth KV object is compiled, linked, or retrieved;
CoMem exposes a single-residual depth axis and quantifies the compute that
remains after selection.

\paragraph{Token selection, compression, and external memory.}
ILRe, REFORM, and GemFilter use intermediate layers to select or gather tokens
before later computation, but do not persist the selected hidden states across
queries~\citep{ilre,reform,gemfilter}. KV-compression methods reduce retained
state after processing~\citep{streamingllm,h2o,snapkv,pyramidkv,minicache}.
MemoryLLM, LongMem, and XC-Cache add latent pools or auxiliary readers
~\citep{memoryllm,longmem,xccache}. CoMem addresses a separate repeated-query
boundary: prepay a chosen number of native decoder layers once, retrieve a
bounded residual pack, and resume the unmodified suffix through a small
adapter.

\paragraph{Positioning.}
The nearest systems differ in backbone, training, tasks, storage tier, and
timing boundary, precluding a numerical cross-paper ranking. The defensible
novelty is narrower and structural: CoMem makes transformer split depth the
controlled variable of reusable-context serving, provides an identical-pack
$j{=}0$ endpoint, and measures the resulting residual object across quality,
latency, storage, Write amortization, and context dependence.

\section{Why Reuse Model Depth?}
\label{sec:motivation}

\begin{figure}[t]
\centering
\includegraphics[width=\columnwidth]{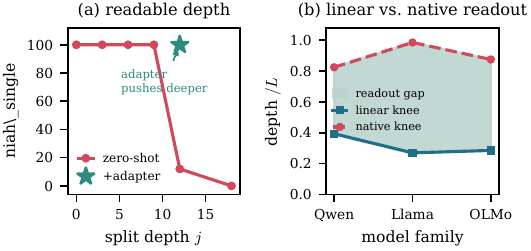}
\caption{\textbf{Intermediate accessibility motivates, but does not validate,
depth reuse.} Frozen probes become informative before native LM-head readout,
but both measures are protocol-dependent; definitions and controls are in
Appendix~\ref{app:ablations}.}
\label{fig:depth}
\end{figure}

The interface does not require a layer at which a model has ``understood'' the
document. It requires only that intermediate states retain useful information
and that later layers can be adapted to read states produced without the query.
Figure~\ref{fig:depth} supplies motivation, while the matched experiment in
Section~\ref{sec:experiments} supplies the actual evidence.

A frozen interface deteriorates as the split moves deeper. With separately
distilled adapters, RULER changes from 98.29 at $j{=}6$ to 96.07 at $j{=}12$
and 55.41 at $j{=}18$, while fixed-pack Read speedup rises from
$1.166\times$ to $1.403\times$ and $1.807\times$
(Appendix Table~\ref{tab:distilled-depth-curve}). Each split uses a separately distilled suffix adapter, yielding a practical
\emph{multi-depth deployment curve}: greater prepaid depth systematically
reduces Read time until interface quality collapses at $j{=}18$. Because adapter
spans and parameter counts change with $j$ and the retained $j{=}12$ Write value
is unavailable, the clean causal estimate remains the matched $j{=}0$ versus
$j{=}12$ endpoint.

Selection and representation also fail separately. BM25 retains lexical
single-needle evidence, but on LongEval the distilled $j{=}12$ reader reaches
only 64--76\% accuracy conditional on a retrieval hit; BABILong exhibits both
misses and hit-conditional readout failures. We therefore evaluate depth under
a fixed evidence pack and diagnose Write context only after retrieval is held
constant.

\section{CoMem}
\label{sec:method}

\paragraph{Interface.}
Let a decoder have $L$ layers and width $d$, with half-open layer ranges and
$h_j$ denoting the input to block $j$. \textsc{Write} divides a document into
$c$-token chunks, runs each through $[0{:}j)$ with positions restarted at zero,
and stores $h_j\in\mathbb{R}^{c\times d}$. \textsc{Select} returns top-$k$
chunk IDs. \textsc{Read} orders them by document position, assigns fresh
contiguous positions to $[\mathrm{BOS};\,\{h_j\};\,h_j(q)]$, and runs
$[j{:}L)$ with full causal cross-pack attention. For $j{=}0$, the same path
stores token IDs and executes all layers, giving the matched endpoint of the
depth axis.

\paragraph{Storage and bounded model work.}
With head width $d_{\mathrm{head}}$, query-head count $n_{\mathrm q}$, and
KV-head count $n_{\mathrm{kv}}$, residual-to-full-KV bytes/token at a common
dtype are
\begin{equation}
\frac{\lvert h_j\rvert}{\lvert\mathrm{KV}\rvert}
=\frac{d}{2L n_{\mathrm{kv}}d_{\mathrm{head}}}
=\frac{n_{\mathrm{q}}}{2L n_{\mathrm{kv}}},
\label{eq:store}
\end{equation}
using $d=n_{\mathrm q}d_{\mathrm{head}}$. For Qwen3-8B this is $1/18$:
8\,KiB rather than 144\,KiB/token in bf16. It is about 1\,GiB at 128k
tokens---far larger than text, so reuse is essential. The model-side Read length
is $\mathrm{sink}+kc+\mathrm{query}$ (nominally 6,657 tokens for
$k{=}12,c{=}512$), independent of stored-context length. The persistent store,
index, and selector are not bounded.

\paragraph{Overlap-Write.}
Chunk-local writes omit lower-layer left context. For the diagnostic variant,
chunk $i$ is prefixed by its preceding $w$ tokens during Write, but only the
current $c$ residuals are stored. Thus persistent bytes, selected-pack length,
and per-query Read/decode are unchanged; Write work and edit invalidation grow.
The overlap sweep tests how much local context is sufficient to repair independently written states; the default full-suite results retain $w{=}0$.

\paragraph{Self-distillation.}
Following knowledge distillation and low-rank adaptation
~\citep{distillation,lora}, a $j{=}12$ student matches a frozen $j{=}0$ teacher
on the teacher's top-64 logit support at every query token. The teacher runs the
same chunk-local pack from $j{=}0$ with the adapter disabled; the student writes
the sink, each context chunk, and the query independently to $h_{12}$, then
resumes the packed query through layers 12--35. If $S_t$ is the teacher's
top-64 vocabulary set at query token $t$, then
$p_t=\mathrm{softmax}(z^T_{t,S_t})$ and
$q_t=\mathrm{softmax}(z^S_{t,S_t})$: both distributions are renormalized only
over the same teacher-selected support, and all logits outside $S_t$ are
discarded by this approximate objective. We did not retain the teacher mass
captured by $S_t$. The token-mean loss is
\begin{equation}
\mathcal L_{\mathrm{distill}}
=0.6\,\mathrm{KL}(p\Vert q)+0.4\,\mathrm{KL}(q\Vert p),
\label{eq:distill}
\end{equation}
with no cross-entropy term. The rank-32 LoRA ($\alpha{=}64$, zero dropout)
updates attention and MLP projections in layers 12--35; the Qwen3-8B backbone
stays frozen. Training uses 4,000 bf16 DDP steps on 2,048-token PG-19 windows
~\citep{pg19}, without benchmark supervision, retrieval, or synthetic
long-context data. Full optimizer, masks, reduction, and artifact details are
in the appendix.

\paragraph{Selection and generation.}
The flagship selector is deterministic token-ID BM25~\citep{bm25}, chosen to
avoid a separately trained retriever. It uses top-12, hop width four, and three
frontier-expansion rounds: round 1 scores the question; later rounds concatenate
tokens from newly selected chunks, score unselected chunks, and break ties by
document index. Frozen dense retrieval is reported only as a diagnostic. During
generation, lower layers cache the query/generation prefix and upper layers
cache the selected pack; the retrieved pack is not rerun for each output token.
Exact masks, positions, and timing boundaries appear in the appendix.

\section{Experiments}
\label{sec:experiments}

\paragraph{Setup.}
The principal model is frozen Qwen3-8B~\citep{qwen3}, with $j{=}12$,
512-token chunks, iterative BM25 top-12, a BOS sink, and the rank-32 LoRA.
Prompting is plain text without chat templates. We evaluate RULER
~\citep{ruler}, BABILong~\citep{babilong}, LongEval~\citep{longchat}, six
LongBench QA datasets~\citep{longbench}, and all 1,986 LoCoMo questions
~\citep{locomo}. Appendices~\ref{app:tasks} and~\ref{app:statistics} give
complete cells, prompts, sample counts, uncertainty, and integrity checks. We keep four
timing boundaries separate: selected-pack model Read, store-ready online
prefill, Write-inclusive pipeline, and external index/store cost.

\subsection{Matched depth reuse}
\begin{table}[!t]
\centering
\small
\setlength{\tabcolsep}{3pt}
\renewcommand{\arraystretch}{1.05}
\begin{tabular}{@{}p{0.24\textwidth}p{0.13\textwidth}p{0.10\textwidth}rrrr@{}}
\toprule
\textbf{Operating point} & \textbf{Stored object} & \textbf{Online depth} &
\textbf{RULER}$\uparrow$ & \textbf{Read} & \textbf{Write} &
\textbf{Store/token} \\
\midrule
Matched raw-text replay ($j{=}0$) & token IDs & 36 layers &
99.19 & 931.9\,ms & per query & $\sim$4--8\,B \\
CoMem + LoRA ($j{=}12$) & $h_{12}$ & 24 layers &
96.07 & 664.4\,ms & once & 8,192\,B \\
Continuous-prefix oracle & $h_{12}$ & 24 layers &
99.19 & --- & per query & 8,192\,B \\
\bottomrule
\end{tabular}
\caption{\textbf{Matched depth reuse delivers $1.403\times$ faster Read at a
3.12-point quality cost.}
All rows use the same paired 15-cell RULER-B cohort ($n{=}1{,}500$), selected
pack, order, sink, mask, examples, and LoRA. The $j{=}0$--$j{=}12$ gap has
paired-bootstrap 95\% CI $[2.36,3.93]$. Read latency uses an approximately
6.5k-token pack from 16k source contexts and three process medians; selection,
persistent I/O, reusable Write, and decode are excluded. The
continuous-prefix oracle computes lower-layer states jointly with the selected
pack and query, exactly recovering 99.19; it therefore shows that the upper
24 layers can reproduce full replay when supplied compatible $h_{12}$ states.}
\label{tab:core-tradeoff}
\end{table}

Table~\ref{tab:core-tradeoff} is the central comparison. Raw-text replay
($j{=}0$) and CoMem receive the same selected chunks in the same order and use
the same sink, mask, examples, and LoRA. Skipping the lower 12 layers reduces
fixed-pack Read from 931.9 to 664.4\,ms ($1.403\times$), but lowers the RULER
macro from 99.19 to 96.07. This gives the matched depth frontier: $1.403\times$ faster Read for a
3.12-point quality cost. The continuous-prefix control exactly matches $j{=}0$,
showing that the upper 24 layers retain sufficient capacity to reproduce replay
when supplied compatible states. It therefore turns the observed loss into a
Write-interface problem that the following controls can diagnose.

Natural-task transfer is uneven. Relative to matched $j{=}0$, $j{=}12$
falls sharply on LongEval (97.2 to 69.0), changes modestly on LoCoMo (41.59 to
38.27), and is nearly unchanged on six LongBench QA datasets (12.31 to 12.15);
BABILong varies by task
(Table~\ref{tab:babilong}). The detailed tables
(Tables~\ref{tab:longeval}, \ref{tab:longbench}, and \ref{tab:locomo}) include
the corresponding measured $j{=}0$ reference cells rather than repeating an
overview macro in each length or dataset column. They support transfer of the
interface, not uniform preservation of task quality.
Data-overlap and judge qualifications appear in the appendix and Limitations.
The external references also expose important boundaries rather than uniform
dominance. In the full-prompt YaRN control, KV-Direct reaches 100 on 128k
single-needle and 89 on multikey versus CoMem's 98/93, while CoMem is much
stronger on variable tracking (99.0 versus 57.8). At 8k--32k, SnapKV and
PyramidKV reach a 99.96 RULER macro after full-prompt prefill; their apparent
72.3 native 15-cell macro is lowered by left-truncated 64k/128k inputs. CoMem's
advantage is therefore bounded read without full-prompt prefill and stronger
long-range tracking, not universal per-cell accuracy.

\subsection{Same-backbone chunk-KV comparison}

Table~\ref{tab:cacheblend-baseline} compares CoMem with a minimal faithful
CacheBlend-style arm~\citep{cacheblend}. The measured dimensions held common
are Qwen3-8B, iterative-BM25 top-12 retrieval, 512-token chunks, plain-text
prompts, and scorers. Both use the same pack-construction policy, but the
retained artifact does not prove identical examples or packs; their length
supports also differ, and only CoMem is adapted. The baseline prefills
full-depth per-chunk KV, reindexes
RoPE when chunks are assembled, and selectively recomputes the highest-deviation
context tokens; a real-model $r{=}1$ self-test agrees with vanilla full prefill
to $2.8\!\times\!10^{-5}$ maximum logit difference and 100\% top-1 agreement.
Increasing $r$ from 0 to .18 improves its 12-cell RULER macro from 67.80 to
74.70, but leaves BABILong qa5 near 49 and LoCoMo substring accuracy near 17.
CoMem separately reports 97.05 on 15 RULER cells, 68.67 on seven BABILong qa5
cells, and 23.36 on the full LoCoMo set. Its single residual
uses 8\,KiB/token rather than 144\,KiB/token for full-depth bf16 GQA KV. Thus,
as a protocol-aligned diagnostic rather than a matched estimate, full-layer KV
still underperforms the residual-plus-upper-recompute path. This is not a native
CacheBlend reproduction or an adaptation-matched causal estimate.

\subsection{Amortization and equal-latency boundaries}

The residual store costs 8,192 bytes/token and must first be written. In the
write-once serving harness, CPU-pinned break-even is 8.9--10.9 queries at 32k
for generated lengths up to 128 tokens; the full GPU/CPU grid is 5.5--10.9.
The 512-token 32k CPU case moves to 94 queries because decode dominates, and
the 32k GPU cell has no finite crossover at that output length. At 128k,
break-even is 25.8--27.2 queries for GPU placement and 27.6--37.4 for CPU
placement through $G{=}128$; at 1M tokens it rises to roughly 180--421 queries.
Each value follows the displayed measured-time equation and includes fetch,
model Read, and decode; for one-off or weakly reused documents, raw text is
cheaper.

We then map the boundary at which depth reuse remains attractive. A latency-only
calibration freezes maximum budgets of 10 replay chunks and 12 CoMem chunks;
selectors may under-fill. With the same iterative-BM25 ranking, raw replay leads
by 11.56 points, establishing that the saved depth does not compensate for its
representation cost on this tested mixture. Replacing only the replay selector
with frozen BGE changes the point difference to $-1.00$, while the unchanged
CoMem arm remains at 53.22. A hierarchical bootstrap that first resamples cells gives
$[-18.67,-5.11]$ for BM25 and $[-10.67,8.33]$ for BGE; leave-one-cell-out
ranges are $[-13.13,-9.50]$ and $[-3.50,1.75]$. The applicable regime is
therefore explicit: BM25 replay has a robust advantage, whereas the
cross-selector BGE diagnostic is unresolved under task-cell dependence. Depth
reuse can buy a larger evidence budget, but whether that budget closes the
representation gap depends strongly on retrieval quality.
Appendix Table~\ref{tab:equal-latency-protocol} exposes the cohort, weights,
generation limits, split, absolute latencies, and bootstrap unit.

\subsection{Write-interface diagnosis and repair}
\begin{table}[!htbp]
\centering
\small
\setlength{\tabcolsep}{3pt}
\begin{tabular}{@{}lcc@{}}
\toprule
\textbf{Lower-layer Write scope} & \textbf{local Read pos.} &
\textbf{document-origin Read pos.} \\
\midrule
Chunk local & 92.5 & 88.0 \\
Document contextual & 100.0 & 100.0 \\
\bottomrule
\end{tabular}
\caption{\textbf{Bounded context-scope $\times$ Read-position diagnosis.}
Paired Qwen3-8B multikey macro over 8k/16k ($n{=}200$), with identical
retrieval, LoRA, and upper-layer Read. Document context adds 7.5 points with
local Read positions and 12.0 with document-origin positions; changing position
alone lowers the chunk-local writer by 4.5 points. The $+4.5$ interaction means
the factors are not additive. This attribution is limited to the displayed
cohort.}
\label{tab:context-position}
\end{table}

Holding retrieval, adapter, and upper-layer Read fixed, the $2{\times}2$
diagnostic separates lower-layer attention scope from Read positions.
Document-contextual Write closes the gap under either position convention;
document-origin positions alone make chunk-local states 4.5 points worse. The $+4.5$ interaction precludes a purely additive explanation and identifies
the Write representation---especially missing lower-layer document context---as
the dominant tested source of the gap rather than position remapping.

Overlap-Write turns this control into a local deployable intervention:
$w{=}32$ raises 92.5 to 98.5 and $w{=}128$ to 99.0 while preserving stored
bytes and per-query work. Its measured wall-clock overhead exceeds the theoretical FLOP ratios, but the
result demonstrates that most of the local gap can be repaired with only 32
left-context tokens and no increase in persistent bytes or online Read. A
separately trained lower-layer Write adapter reaches the same 98.5 on this
cohort; details and paired tests are in Table~\ref{tab:write-context}. Both
repairs remain synthetic diagnostics rather than natural-task or end-to-end
validation.

\subsection{Efficiency and timing boundaries}
The main store-ready reference is the H20 Pareto harness: stock Qwen3-8B
without the CoMem adapter prefills the full 128k source in 50.59\,s, whereas
adapted CoMem processes its selected pack in 1.32\,s
(Table~\ref{tab:online-prefill}). This composed operating point combines bounded
selection with depth reuse and excludes one-time document Write and
external-store fetch; because its dense arm is LoRA-off, it is not a
same-adapter comparison. Finally, the same-adapter L20A pipeline comparison processes the full source in
6.035\,s and CoMem in 2.202\,s ($2.74\times$), with the latter including
document Write, CPU BM25 retrieval, query Write, and upper-layer Read but
excluding BM25-index construction and external I/O (Table~\ref{tab:eff}). The
two measurements therefore separate store-ready online work from a
Write-inclusive same-adapter pipeline rather than forming interchangeable
speedup denominators; the appendix retains the separate H20 cohort for
completeness.
The Write-inclusive comparison is length-dependent: CoMem is slower at 8k and
16k (0.53/0.81$\times$) and crosses dense prefill only from 32k onward, reaching
2.74$\times$ at 128k.

On the fixed-pack matched harness, the 267.5\,ms Read saving becomes only
$1.07$--$1.09\times$ end-to-end once 2.76--2.86\,s autoregressive decode is
included. This is the expected decode bottleneck for bounded-read systems;
speculative decoding and prefix caching are orthogonal and not measured here.

\FloatBarrier
\section{Conclusion}
\label{sec:conclusion}

CoMem exposes prepaid depth as a reusable-context axis. Prepaying 12 Qwen3-8B
layers gives a $1.403\times$ Read speedup at a 3.12-point RULER cost; matched
replay and Write controls delimit the trade-off.

\section*{Limitations}

CoMem is evaluated mainly on one Qwen3-8B model, a lexical selector, and English
benchmarks. Cross-scale and sparse-MoE results test implementation portability,
not matched replication, and some large-model cells are smaller. The flagship
adapter is one batch-8 run; two matched-data runs use effective batch 3, so they
conflate seed and optimization noise. We still lack same-backbone,
same-hardware implementations of the broader PIC and learned modular-cache
families. The included CacheBlend-style experiment is a minimal faithful
full-depth chunk-KV control with global RoPE reindexing and selective
recomputation, not the complete native CacheBlend scheduler, cache manager, or
production serving stack; it is training-free whereas CoMem uses a distilled
adapter. Published systems still differ in model, task, training, storage tier,
and timing boundary, so Table~\ref{tab:priorart} remains structural rather than
a cross-paper ranking. The controlled depth evidence is the
$j{=}0\!\rightarrow\!12$ endpoint; separately trained splits form a deployment
curve, not a compute-matched causal frontier.

The workload assumes repeated queries over a stable corpus. Residual storage is
linear and much larger than text, selector/store latency grows with corpus size,
and model-specific states generally require rewriting after changes to the
backbone, tokenizer, split, adapter, or lower-layer implementation. One-off,
low-reuse, or frequently edited documents may favor raw text; Overlap-Write can
also invalidate a neighboring chunk after an edit. Latency and break-even depend
on hardware, kernels, batch size, storage placement, and timing boundaries. The
large-output crossover can be unstable when decode leaves only a small
per-query latency difference; serving reports single-query medians rather than
concurrent throughput or p95/p99 tails. Quantization,
eviction, update contention, and production multi-tenancy are not evaluated.
The 50.59/1.32\,s and 6.035/2.202\,s efficiency figures come from separate H20
and L20A harnesses with different adapter and inclusion boundaries; they are
labeled separately and must not be combined into one speedup.

Quality remains task-, selector-, and evidence-budget-dependent. Top-$k$ limits
coverage; lexical retrieval transfers poorly to low-overlap and non-whitespace
settings; and residual readout can fail after a retrieval hit. Overlap-Write is
validated only on paired synthetic multikey instances, without a natural-task or
Write-inclusive repaired frontier. The trained lower-layer Write adapter is
evaluated on the same synthetic cohort and therefore remains a mechanism
upper-bound rather than evidence of natural-task repair. Beyond-native-window results are positional
stress tests. In the equal-latency mixture, BM25 replay wins robustly, while the
frozen-BGE difference spans both signs; cells are equally weighted and the 100
LoCoMo items come from one conversation, preventing conversation-cluster
inference. Thus this diagnostic does not establish a selector-independent
advantage.

LoCoMo uses a mutable, undated \texttt{gpt-4o} judge; saved decisions and a
200-item DeepSeek-V3 audit preserve ordering but not exact reproducibility under
a fixed snapshot. PG-19 overlaps InfiniteBench long-book support, so that
comparison was removed. A later audit localizes LongBench overlap to
NarrativeQA and finds at most a 0.13-point macro-F1 change after strict
clean-subset recomputation; LoCoMo is clean and LongEval is synthetic by
construction. This supports robustness to detected near-verbatim overlap, not a
guarantee against semantic paraphrase or exposure in other pretraining corpora.
Finally, probes measure
readout compatibility rather than the location or causal necessity of
``understanding'' or factual knowledge. Long-form, code, multilingual,
multimodal, and dynamic-memory use remain open.

\section*{Ethical Considerations}
CoMem inherits the underlying model's hallucination, bias, unsafe-generation,
and misuse risks; cheaper access to long histories can scale both beneficial and
harmful applications. Retrieved or cached material may disclose sensitive or
unauthorized text, and residual tensors may permit inversion or membership
inference, although we do not test these attacks. Deployments should protect the
store at least as strongly as source text through authorization, tenant
isolation, encryption, auditing, pre-Write filtering/redaction, output
monitoring, and verifiable deletion; large persistent stores and Overlap-Write
also require careful update invalidation. Bounded reads reduce per-query compute,
but training and evaluation still consume energy; measured final-run hardware
and compute appear in Appendix~\ref{app:reproducibility}.

We collect no new human-subject data or annotations. Experiments use released
models, corpora, and benchmarks; LoCoMo conversations are model-generated. The
anonymous artifact excludes model weights, benchmark text, credentials, private
API metadata, and raw private responses. Subject to upstream terms, it provides
the adapter, evaluation code, permissible prediction artifacts or hashes, judge
prompts/parsed decisions, timing records, licenses, and provenance needed to
audit the reported results.

\bibliography{qcmem}
\bibliographystyle{colm2026_conference}

\clearpage
\appendix
\raggedbottom
\section{Additional Results and Ablations}
\label{sec:appendix}

This appendix contains the detailed task tables, mechanism ablations, and
cross-scale generality results summarized in the main paper.
Appendix~\ref{app:statistics} separately documents aggregation, uncertainty,
and the LoCoMo judge protocol.

\subsection{Systems accounting and equal-latency audit}
\label{app:systems}

\paragraph{Detailed interface positioning.}
Text RAG is the matched $j{=}0$ endpoint: it stores token IDs and reruns every
layer after selection. Reusable-representation methods fix a compressed or
intermediate interface and adapt the downstream computation. Semantic caches
match similar inputs against per-layer activation banks, whereas hidden-state
restoration reconstructs standard layer-wise KV. Chunk-KV and PIC systems store
full-depth KV objects and vary linking or repair; learned modular KV methods
move more document computation into training or compilation. CoMem instead
stores one document residual at a chosen split and directly executes the native
suffix. These families differ in backbone, training, storage tier, and timing
boundary, so Table~\ref{tab:priorart} is structural rather than a numerical
ranking; Table~\ref{tab:cacheblend-baseline} provides the focused same-backbone
diagnostic.

\paragraph{Secondary dense-context operating point.}
When the persistent store already exists, the select-first path reads a fixed
6,657-token pack at every source length. Table~\ref{tab:online-prefill} reports
1.32\,s and 18.7\,GB at 128k versus 50.59\,s and 50.0\,GB for stock dense
prefill on the same H20. The resulting 38.3$\times$ combines bounded selection
with depth reuse; it is neither the incremental depth effect nor an end-to-end
repeated-query comparison.

\begin{center}
\captionsetup{hypcap=false}
\centering
\small
\setlength{\tabcolsep}{4.5pt}
\begin{tabular}{@{}lrrrrr@{}}
\toprule
\textbf{Length} & \shortstack{\textbf{Dense}\\\textbf{prefill}} & \shortstack{\textbf{CoMem}\\\textbf{prefill}} & \textbf{Speedup} & \shortstack{\textbf{Dense}\\\textbf{peak}} & \shortstack{\textbf{CoMem}\\\textbf{peak}} \\
\midrule
8k   & 1.10\,s  & 0.94\,s & 1.17$\times$ & 18.8\,GB & 18.6\,GB \\
32k  & 5.94\,s  & 1.09\,s & 5.45$\times$ & 25.0\,GB & 18.7\,GB \\
128k & 50.59\,s & 1.32\,s & 38.3$\times$ & 50.0\,GB & 18.7\,GB \\
\bottomrule
\end{tabular}
\captionof{table}{\textbf{Store-ready online-prefill operating points.}
Measurements use one H20, Qwen3-8B, bf16/SDPA, and the median of three runs
after one warmup. Dense is the stock, LoRA-off Qwen3 path and processes the
complete source; CoMem reads a fixed
6,657-token top-12 pack with its rank-32 LoRA. The persistent store is already
built, so these values isolate the store-ready online path and exclude one-time
Write and external-store fetch. The matched incremental depth effect is the
separate $1.403\times$ result in Table~\ref{tab:core-tradeoff}; the
Write-inclusive same-adapter L20A cohort appears in Table~\ref{tab:eff}.
The 50.59\,s dense row is therefore a stock-dense master reference, not a
same-adapter control.}
\label{tab:online-prefill}
\end{center}

\begin{table}[!htbp]
\centering
\small
\setlength{\tabcolsep}{5pt}
\begin{tabular}{@{}lrrrr@{}}
\toprule
\textbf{Store} & \textbf{$G{=}1$} & \textbf{$G{=}32$} &
\textbf{$G{=}128$} & \textbf{$G{=}512$} \\
\midrule
32k, CPU-pinned & 8.9 & 9.2 & 10.9 & 94.0 \\
32k, GPU-resident & 8.4 & 7.7 & 5.5 & $\infty$ \\
128k, CPU-pinned & 27.6 & 29.7 & 37.4 & 520.1 \\
128k, GPU-resident & 25.8 & 26.8 & 27.2 & 164.2 \\
1M, CPU-pinned & 198.2 & 266.7 & 421.3 & 302.6 \\
1M, GPU-resident & 180.2 & 183.0 & 190.6 & 574.7 \\
\bottomrule
\end{tabular}
\caption{\textbf{Measured repeated-query break-even $Q^\star$ over the full
serving grid.}
$G$ is generated tokens; a cell is the number of queries after which cumulative
CoMem time is lower than matched $j{=}0$ raw-text replay. Both arms share
selection, example, and pack; selection cancels in
$Q^\star=(W_{\mathrm{CoMem}}-I_{j=0})/
(T_{j=0}-T_{\mathrm{CoMem}})$. The 32k CPU cell uses
$W_{\mathrm{CoMem}}{=}2.253$\,s and a 2.16\,ms raw-token index build.
Values derive from three process medians and include store fetch, model Read,
and decode. The 18 per-process records and aggregate are released with
SHA-256 manifests. $\infty$ means CoMem's measured per-query time is not
lower, so no finite query count amortizes its one-time Write in that cell.
Large-$G$ values are sensitive because decode leaves a small per-query
denominator; they are workload-specific measurements rather than a monotonic
scaling law.}
\label{tab:serving-crossover}
\end{table}

\begin{table}[!htbp]
\centering
\small
\setlength{\tabcolsep}{3.2pt}
\begin{tabular}{@{}lrrrr@{}}
\toprule
\textbf{Method / recompute} & \textbf{RULER} & \textbf{BABI-qa5} &
\textbf{LoCoMo} & \textbf{Store/tok.} \\
\midrule
CacheBlend-style, $r{=}0$    & 67.80 & 49.33 & 17.07 & 144\,KiB \\
CacheBlend-style, $r{=}.10$  & 72.57 & 48.50 & 16.77 & 144\,KiB \\
CacheBlend-style, $r{=}.15$  & 73.78 & 49.83 & 17.07 & 144\,KiB \\
CacheBlend-style, $r{=}.18$  & 74.70 & 49.17 & 17.37 & 144\,KiB \\
\midrule
CoMem ($j{=}12$) & 97.05 & 68.67 & 23.36 & 8\,KiB \\
\bottomrule
\end{tabular}
\caption{\textbf{Same-backbone full-depth chunk-KV comparison.}
The baseline uses independently prefetched 36-layer KV, global RoPE reindexing,
and selective recomputation; its $r{=}1$ self-test matches full prefill. Both
arms use Qwen3-8B, 512-token chunks, iterative-BM25 top-12, plain text, and the
same scorers. However, CacheBlend reports 12 RULER cells at 4k--32k and six
BABI-qa5 cells at 0k--16k, whereas the CoMem entries summarize 15 cells at
8k--128k and seven cells at 0k--32k. LoCoMo has 1,986 questions for both, but
the artifact does not prove itemwise pack identity. The training-free baseline
and adapted CoMem are therefore protocol-aligned diagnostics, not paired or
adaptation-matched estimates. Full-depth bf16 GQA KV costs 147,456 bytes/token
versus 8,192 for CoMem. The recorded BABILong \texttt{resume\_j=6} is a host
default; CacheBlend executes all 36 layers.}
\label{tab:cacheblend-baseline}
\end{table}

\paragraph{Equal-latency audit.}
\begin{table}[!htbp]
\centering
\small
\setlength{\tabcolsep}{2.2pt}
\resizebox{\columnwidth}{!}{%
\begin{tabular}{@{}lrrrrr@{}}
\toprule
\textbf{Replay selector} & \textbf{$k_{\max}$} &
\textbf{Replay / CoMem} & \textbf{$\Delta$} &
\textbf{Hier. 95\% CI} & \textbf{LOCO range} \\
\midrule
Iterative BM25 & 10 & 64.78 / 53.22 & $-11.56$ &
$[-18.67,-5.11]$ & $[-13.13,-9.50]$ \\
Frozen BGE & 10 & 54.22 / 53.22 & $-1.00$ &
$[-10.67,8.33]$ & $[-3.50,1.75]$ \\
\bottomrule
\end{tabular}%
}
\caption{\textbf{Equal-latency quality exposes selector choice as a first-order
frontier variable.}
A latency-only calibration fixes replay budget 10 against CoMem budget 12
within a predeclared $\pm5\%$ band. Quality is the equal-weight mean of nine
100-example task--length cells; $\Delta$ is CoMem minus replay. The BM25 replay
advantage is robust to hierarchical and leave-one-cell-out analyses. The
frozen-BGE point estimate favors replay by one point, but both dependence-aware
ranges span zero. Appendix Table~\ref{tab:equal-latency-protocol} gives the
complete protocol and sensitivity analyses.}
\label{tab:equal-latency}
\end{table}

Table~\ref{tab:equal-latency-protocol} reconstructs the complete saved protocol
behind Table~\ref{tab:equal-latency}, including the nine cells, equal cell
weights, disjoint latency calibration, generation limits, absolute TTFT,
hardware, storage assumptions, all cell deltas, dependence-aware bootstraps,
leave-one-cell-out sensitivity, and the original pooled-IID interval.

\begin{table}[!htbp]
\centering
\small
\setlength{\tabcolsep}{3pt}
\begin{tabular}{@{}p{0.20\textwidth}p{0.25\textwidth}p{0.48\textwidth}@{}}
\toprule
\textbf{Protocol item} & \textbf{Setting} & \textbf{Auditable detail} \\
\midrule
Quality cohort & 9 cells, $n{=}100$/cell &
BABILong qa1/qa2 at 4k and 16k; LongEval at 8k and 16k;
RULER \texttt{niah\_multikey\_1} at 8k and 16k; first 100 LoCoMo
items. Each cell contributes one equally weighted accuracy mean. The retained
LoCoMo slice is entirely conversation 0, so its 100 questions cannot support a
conversation-cluster bootstrap. \\
Generation/scoring & Greedy, no chat &
BABILong: 20 tokens and official \texttt{compare\_answers};
LongEval: 16 tokens and numeric exact match; RULER: 48 tokens and official
\texttt{string\_match\_all}; LoCoMo diagnostic: 48 tokens and deterministic
local substring/F1-threshold or abstention accuracy, not the GPT-4o headline
judge. \\
Calibration split & Latency only, disjoint &
Three RULER multikey 32k documents at reserved indices 900--902
(\texttt{PYTHONHASHSEED=0}, experiment seed 42), disjoint from quality indices
0--99. Integer $k\in\{2,4,6,8,10,12,14,16,20,24\}$ is selected only by TTFT
within $\pm5\%$ of the CoMem budget-12 anchor. Here $k$ is a maximum chunk
budget; iterative BM25 may under-fill. \\
Timing & One NVIDIA H20; bf16/SDPA &
Three independent processes per $k$, five warmups and 20 repetitions each;
reported TTFT is the median of process medians. It includes online selection,
raw/residual fetch, sink/query Write, and model prefill to first logits; decode,
one-time document Write, and index construction are excluded. BM25 replay
budget 10 is 703.7\,ms versus CoMem 698.3\,ms (GPU) / 699.8\,ms (CPU);
BGE replay is 697.8\,ms versus 700.5 / 702.2\,ms. \\
Cell deltas: BM25 &
CoMem(12) minus replay(10), points &
qa1: $-6$ (4k), $-17$ (16k); qa2: $-3$, $-10$;
LongEval: $-26$ (8k), $-28$ (16k); RULER: $-6$, $-9$;
LoCoMo: $+1$. Aggregate $-11.56$. \\
Cell deltas: BGE &
CoMem(12) minus replay(10), points &
qa1: $-5$ (4k), $-21$ (16k); qa2: $0$, $-23$;
LongEval: $0$ (8k), $+16$ (16k); RULER: $+1$, $+19$;
LoCoMo: $+4$. Aggregate $-1.00$. \\
Dependence-aware inference &
100,000 paired bootstrap replicates &
\textbf{Stratified fixed-cell:} keep all nine cells and resample exactly
$n{=}100$ pairs inside each cell; intervals are $[-14.33,-8.78]$ (BM25) and
$[-4.56,2.56]$ (BGE). \textbf{Hierarchical:} resample nine cell labels, then
resample $n{=}100$ pairs within every selected occurrence; intervals are
$[-18.67,-5.11]$ and $[-10.67,8.33]$. \textbf{Leave-one-cell-out:}
$[-13.13,-9.50]$ (all 9 negative) and $[-3.50,1.75]$ (7 negative, 2 positive).
Deterministic base seed is 20260804. \\
IID sensitivity & Original v5 analysis &
Pool all 900 paired differences and resample 900 pairs:
$[-14.44,-8.67]$ (BM25) and $[-4.67,2.67]$ (BGE). These intervals are retained
for continuity but are not the task-mixture inference. \\
\bottomrule
\end{tabular}
\caption{\textbf{Protocol-complete equal-latency audit and dependence-aware
reanalysis.}
Phase B intentionally uses frozen BGE for raw replay and iterative BM25 for
CoMem, so it tests selector sensitivity rather than a same-pack depth effect.
The reanalysis reads only saved per-example scores; it runs no model. The
anonymous release contains score-only exports (no benchmark text, gold answers,
or predictions), the script, result JSON, and SHA-256 manifests.}
\label{tab:equal-latency-protocol}
\end{table}

\paragraph{Content-depth probe protocol.}
For SST2, WiC, and RTE we extract every layer once from a fixed stratified pool
(up to 3,000 train examples), then fit L2 logistic regressions on five independent
60/20/20 train/dev/test splits. Features are standardized; $C\in\{0.1,1,10\}$
is selected on development data. For layer accuracy $a(l)$, the knee is
$\min\{l:a(l)\geq0.98\max_l a(l)\}$. The native readout instead applies the
model's own final norm and LM head to each layer using fixed verbalizers. Lexical
uni/bi-grams, token length, random labels, majority class, and balanced training
serve as controls. On SST2, probe peak is about 0.90 versus 0.71 lexical, 0.56
position/majority, and 0.54 random-label accuracy; WiC/RTE are lower-selectivity
and have wider per-task intervals. The reported confidence intervals are
Student-$t$ intervals over task--split knees, not training-seed uncertainty.

\subsection{Depth, selection, and interface ablations}
\label{app:ablations}

\paragraph{Split depth and self-distillation.}
The probe values plotted in Figure~\ref{fig:depth} use five stratified data
splits for Qwen and OLMo plus a matched-support Llama SST2 result; the protocol
is specified above. Table~\ref{tab:depth-tradeoff} gives the frozen
$j=0/6/9/12$ quality sweep and same-$j$ distilled control;
Table~\ref{tab:distilled-depth-curve} gives separately distilled
$j=6/9/12/18$ deployment points. Table~\ref{tab:adapter} provides a separate
same-$j$ RULER control.
Finally, Table~\ref{tab:hcache-lora} uses an HCache-style checkpoint path only
as an interface-adaptation diagnostic, showing that the readout repair transfers
beyond CoMem.

\begin{table}[!htbp]
\centering
\small
\setlength{\tabcolsep}{3pt}
\resizebox{\columnwidth}{!}{%
\begin{tabular}{@{}lccccc@{}}
\toprule
\textbf{Configuration}
& \textbf{Split $j$}
& \textbf{LoRA}
& \textbf{LoCoMo Judge}
& \multicolumn{2}{c}{\textbf{BABILong mean}} \\
\cmidrule(l){5-6}
& & & & \textbf{qa1 / qa2} & \textbf{qa5} \\
\midrule
Matched raw-text replay
& 0
& no
& \textbf{41.59}
& 66.3 / 38.9
& 63.3 \\

CoMem frozen
& 6
& no
& 32.78
& 44.6 / 22.6
& 53.9 \\

CoMem frozen
& 9
& no
& 29.15
& 42.4 / 19.6
& 55.6 \\

CoMem frozen
& 12
& no
& 24.52
& 33.4 / 18.0
& 60.3 \\

\textbf{CoMem distilled}
& 12
& yes
& 38.27
& 55.6 / 27.0
& \textbf{68.7} \\
\bottomrule
\end{tabular}%
}
\caption{Controlled comparison of retrieval, split depth, and
self-distillation using Qwen3-8B. All configurations follow the same
plain-text prompting protocol, with model-specific chat templating disabled
(\texttt{chat\_template=False}), and retrieve the same top-12 pack.
For the frozen configurations, readout fidelity decreases as the split moves
deeper. At the same split $j{=}12$, the rank-32 LoRA improves the LoCoMo Judge
score by 13.75 points and the BABILong qa1, qa2, and qa5 means by
22.2, 9.0, and 8.4 points, respectively. BABILong entries are rounded task means
from the underlying 21 task--length cells; the raw-cell $j{=}0$ macro is 56.14. The $j{=}0$ configuration performs
full-depth recomputation over the retrieved pack, providing a selected-pack
reference without lower-layer reuse and thus no
depth-based reduction in query-time computation.}
\label{tab:depth-tradeoff}
\end{table}
\begin{table}[!htbp]
\centering
\small
\begin{minipage}[t]{\textwidth}
\centering
\setlength{\tabcolsep}{5pt}
\begin{tabular}{@{}llccc@{}}
\toprule
\textbf{Task} & \textbf{Adapter} & \textbf{8k} & \textbf{16k} & \textbf{32k} \\
\midrule
\multirow{2}{*}{niah\_single\_2}
  & off & 36 & 7 & 22 \\
  & on  & \textbf{100} & \textbf{100} & \textbf{100} \\
\midrule
\multirow{2}{*}{niah\_multikey\_1}
  & off & 8 & 3 & 1 \\
  & on  & \textbf{91} & \textbf{95} & \textbf{92} \\
\bottomrule
\end{tabular}
\captionof{table}{Same-$j{=}12$ RULER adapter ablation ($n{=}100$/cell).
Each on/off pair shares task, length, selector, and cohort.}
\label{tab:adapter}
\end{minipage}

\vspace{6pt}
\begin{minipage}[t]{\textwidth}
\centering
\setlength{\tabcolsep}{6pt}
\begin{tabular}{@{}lcc@{}}
\toprule
\textbf{Path} & \textbf{LoRA} & \textbf{LoCoMo} \\
\midrule
HCache-style ($j{=}12$) & no  & 13.29 \\
HCache-style ($j{=}12$) & yes & \textbf{31.17} \\
\bottomrule
\end{tabular}
\captionof{table}{Retrieval-free HCache-style checkpoint control. The shared
split and harness isolate interface adaptation, showing that the readout repair
transfers beyond CoMem.}
\label{tab:hcache-lora}
\end{minipage}
\end{table}

\begin{table}[!htbp]
\centering
\small
\setlength{\tabcolsep}{5pt}
\textit{Quality}\par\vspace{2pt}
\begin{tabular}{@{}rrrrr@{}}
\toprule
\textbf{Split $j$} & \textbf{LoRA params} & \textbf{RULER B} & \textbf{Full $-$ deployed (95\% CI)} & \textbf{LoCoMo} \\
\midrule
6  & 72.745M & 98.29 & $0.91$ $[0.49,1.39]$ & 40.38 \\
9  & 65.470M & 97.55 & $1.65$ $[1.04,2.31]$ & 39.02 \\
12 & 58.196M & 96.07 & $3.12$ $[2.36,3.93]$ & 38.27 \\
18 & 43.647M & 55.41 & $43.79$ $[41.77,45.85]$ & 28.65 \\
\bottomrule
\end{tabular}

\vspace{5pt}
\textit{Systems}\par\vspace{2pt}
\begin{tabular}{@{}rrrr@{}}
\toprule
\textbf{Split $j$} & \textbf{Read (ms)} & \textbf{Read speedup} & \textbf{Write (ms)} \\
\midrule
6  & 830.3 & $1.166\times$ & 133.4 \\
9  & 748.3 & $1.273\times$ & 196.9 \\
12 & 664.4 & $1.403\times$ & --- \\
18 & 499.5 & $1.807\times$ & 390.3 \\
\bottomrule
\end{tabular}
\caption{Multi-depth deployment curve. Each split has a separately trained rank-32
adapter on layers $[j{:}36)$ under the same data order, seed, steps, optimizer,
and KL recipe. RULER uses Cohort B ($n{=}1{,}500$ paired examples); gap CIs are
paired bootstrap intervals against full replay with that depth's adapter, and
all McNemar tests reject equality ($p\leq3.05\times10^{-5}$). LoCoMo reports the
full-set GPT-4o judge point estimate ($n{=}1{,}986$). Read/Write use a fixed 16k
same-pack input and three process medians; retrieval, I/O, and decode are
excluded, and total latency remains decode-dominated. The matched fixed-pack
Write value was not retained for the earlier $j{=}12$ artifact. Adapter spans
and parameter counts differ, so this is not a training-compute-matched depth
ablation.}
\label{tab:distilled-depth-curve}
\end{table}

\paragraph{Selection.}
Table~\ref{tab:selector} compares oracle, BM25, recency, and reader attention.
BM25 is near oracle on the lexical needle tasks and is more reliable than the
tested training-free alternatives at long lengths. Variable tracking instead
requires iterative expansion along literal variable links; a three-round CPU
microbenchmark costs 188.8\,ms at 128k, versus 41.0\,ms for one-shot BM25, without
an additional model forward. This is a low-overhead selector in the tested
regime, not a constant-time one: lookup grows approximately linearly with store
size (Table~\ref{tab:store-scaling}). A frozen BGE-large-en-v1.5 retriever is
also evaluated on a focused cohort: it preserves strong hit-conditional reader
accuracy but loses recall as context grows. We do not train or tune a retriever,
so our results establish lexical and frozen-dense operating points rather than
selector optimality.

\begin{table}[!htbp]
\centering
\small
\begin{minipage}[t]{\textwidth}
\centering
\setlength{\tabcolsep}{5pt}
\begin{tabular}{@{}llcccc@{}}
\toprule
\textbf{Task} & \textbf{Len} & \textbf{BM25} & \textbf{Recency} & \textbf{ReaderAttn} & \textbf{Oracle} \\
\midrule
\multirow{3}{*}{niah\_single}   & 8k  & 100 & 100 & 100 & 100 \\
                                & 16k & 100 & 100 & 100 & 100 \\
                                & 32k & 100 & 82  & 73  & 100 \\
\midrule
\multirow{3}{*}{niah\_multikey} & 8k  & 99 & 98 & 97 & 100 \\
                                & 16k & 99 & 88 & 90 & 100 \\
                                & 32k & 99 & 54 & 60 & 100 \\
\midrule
\multirow{3}{*}{var-track}      & 8k  & 99.4 & 99.2 & 99.8 & N/A \\
                                & 16k & 92.6 & 92.4 & 92.4 & N/A \\
                                & 32k & 32.0 & 41.2 & 27.8 & N/A \\
\bottomrule
\end{tabular}
\captionof{table}{Single-pass RULER selector sweep ($n{=}100$; no chat).
Entries are the peak over $k\in\{4,8,12,16,24\}$. Oracle confirms that the
needle-task gap can be selection-limited; it is undefined for variable tracking.}
\label{tab:selector}
\end{minipage}

\vspace{6pt}
\begin{minipage}[t]{\textwidth}
\centering
\setlength{\tabcolsep}{5pt}
\begin{tabular}{@{}lccccc@{}}
\toprule
\textbf{Measure} & \textbf{qa1} & \textbf{qa2} & \textbf{LE} & \textbf{MK} & \textbf{LoCoMo} \\
\midrule
Dense recall@12 & .52 & .45 & .58 & .81 & .70 \\
Raw reader quality & .45 & .31 & .57 & .82 & .18 \\
\bottomrule
\end{tabular}
\captionof{table}{Frozen BGE-large-en-v1.5 diagnostic at 16k (LoCoMo at native
length), with the same chunks and a Qwen3-8B $j{=}0$ reader ($n{=}100$/cell).
Long-context recall can be the bottleneck even when hit-conditional reading is
strong. LE/MK denote LongEval/RULER multikey.}
\label{tab:dense-retriever}
\end{minipage}
\end{table}

\paragraph{Write-context factorization.}
The complete $2{\times}2$ control is reported in the main paper
(Table~\ref{tab:context-position}). It confirms that the position-only
intervention does not repair chunk-local states and that context and position
interact. The full overlap and trained-Write sweep is reported in
Table~\ref{tab:write-context}.

\begin{table}[!htbp]
\centering
\small
\setlength{\tabcolsep}{4pt}
\begin{tabular}{@{}lccc@{}}
\toprule
\textbf{Writer} & \textbf{Context} & \textbf{Macro}$\uparrow$ &
\textbf{Write FLOPs} \\
\midrule
Chunk local ($w{=}0$) & current chunk & 92.5 & $1.000\times$ \\
Overlap ($w{=}32$) & 32-token left prefix & 98.5 & $1.057\times$ \\
Overlap ($w{=}64$) & 64-token left prefix & 98.5 & $1.115\times$ \\
Overlap ($w{=}128$) & 128-token left prefix & \textbf{99.0} & $1.229\times$ \\
Trained chunk-local Write$^\dagger$ & current chunk & 98.5 & $1.000\times$ \\
Document-context oracle & full prefix & 100.0 & $O(N)$ forward \\
\bottomrule
\end{tabular}
\caption{\textbf{A 32-token Write overlap recovers 6.0 of the 7.5-point local
context gap.}
Paired Qwen3-8B results on RULER \texttt{niah\_multikey\_1} at 8k and 16k
($n{=}200$ pooled). For $w{=}32$, the gain over chunk-local Write has 95\% CI
$[3.0,9.5]$. Every deployable overlap row preserves persistent storage and
per-query Read/decode; only one-time Write work and edit invalidation increase.
FLOPs are theoretical lower-12 Write ratios, while measured wall-clock overhead
is larger because execution remains chunk-level.
$^\dagger$A separately trained lower-layer Write LoRA (layers 0--11) reaches
98.5 at step 2000; its gain over chunk-local Write is significant
($p{=}5{\times}10^{-4}$, McNemar), while its residual gap to the
document-context oracle is not ($p{=}0.25$). This uses the same paired synthetic
cohort and is an upper-bound diagnostic, not a natural-task validation.}
\label{tab:write-context}
\end{table}

\paragraph{Read interaction and chunk size.}
Table~\ref{tab:crosschunk} compares full cross-chunk attention with
block-diagonal reuse. Table~\ref{tab:chunk} reports the chunk-size sweep; its
entries are point estimates rather than a multi-seed stability ranking.

\begin{table}[!htbp]
\centering
\small
\begin{minipage}[t]{\textwidth}
\centering
\setlength{\tabcolsep}{4.5pt}
\begin{tabular}{@{}llccccc@{}}
\toprule
\textbf{Method} & \textbf{RULER task} & \textbf{8k} & \textbf{16k} & \textbf{32k} & \textbf{64k} & \textbf{128k} \\
\midrule
\multirow{2}{*}{\textbf{CoMem}} & niah\_single\_2 & 100.0 & 100.0 & 99.0 & 99.0 & 100.0 \\
& var-track & 96.6 & 97.6 & 98.8 & 99.0 & 95.8 \\
\midrule
\multirow{2}{*}{StreamingLLM} & niah\_single & 85.0 & 36.0 & 20.0 & 10.0 & 2.0 \\
& var-track & 41.8 & 2.2 & 0.6 & 0.2 & 0.8 \\
\bottomrule
\end{tabular}
\captionof{table}{Equal-budget, no-chat RULER comparison ($n{=}100$/cell).
CoMem and StreamingLLM use the same nominal budget and scorer but not paired
examples; CoMem retrieves about 6.5k relevant tokens, whereas StreamingLLM keeps
recent tokens.}
\label{tab:slm}
\end{minipage}

\vspace{6pt}
\begin{minipage}[t]{\textwidth}
\centering
\setlength{\tabcolsep}{6pt}
\begin{tabular}{@{}lccc@{}}
\toprule
\textbf{Task} & \textbf{Full attn.} & \textbf{Block-diag.} & \boldmath$\Delta$ \\
\midrule
RULER niah\_single   & 100 / 100 & 100 / 96 & 0 / $+$4 \\
RULER niah\_multikey & \textbf{96 / 94} & 60 / 32 & $+$36 / $+$62 \\
BABILong qa2         & 36 / 20 & 17 / 13 & $+$19 / $+$7 \\
BABILong qa5         & 78 / 69 & 78 / 55 & 0 / $+$14 \\
\bottomrule
\end{tabular}
\captionof{table}{Full cross-chunk attention versus block-diagonal reuse
(8k/16k, no chat). RULER uses $n{=}50$ and BABILong $n{=}100$ with identical
top-12 retrieval. Joint recomputation matters most for multi-fact tasks.}
\label{tab:crosschunk}
\end{minipage}

\vspace{6pt}
\begin{minipage}[t]{\textwidth}
\centering
\setlength{\tabcolsep}{4pt}
\begin{tabular}{@{}lrrcccc@{}}
\toprule
\textbf{chunk} & \textbf{read tokens} & \textbf{uncached step} & \textbf{8k} & \textbf{16k} & \textbf{32k} & \textbf{64k} \\
& & \textbf{(s/token)} & & & & \\
\midrule
128  & 1,665  & 0.19 & 91.0 & 90.0 & 81.0 & 85.0 \\
256  & 3,329  & 0.37 & 80.0 & 90.0 & 90.0 & \textbf{94.0} \\
512  & 6,657  & 0.72 & 89.0 & \textbf{95.0} & \textbf{97.0} & \textbf{94.0} \\
1024 & 13,313 & 1.60 & \textbf{100.0} & 89.0 & 92.0 & \textbf{94.0} \\
\bottomrule
\end{tabular}
\captionof{table}{Chunk-size trade-off on RULER \texttt{niah\_multikey}
($n{=}100$, no chat). Packed-step cost grows approximately linearly with read
length; accuracy values are point estimates, not a multi-seed claim.}
\label{tab:chunk}
\end{minipage}
\end{table}

\subsection{Full chat-template-free benchmark comparisons}
\label{app:tasks}
Tables~\ref{tab:h2h}--\ref{tab:longbench} give the full audited comparisons for
RULER, LongEval, BABILong, and LongBench. Main rows use the unified no-chat
protocol; they include KV-Direct~\citep{kvdirect}, InfLLM~\citep{infllm},
StreamingLLM~\citep{streamingllm}, and MemoryLLM~\citep{memoryllm}. MemoryLLM
uses its released Llama-3 checkpoint~\citep{llama3}, while the marked LLoCO
row uses its released native setup~\citep{lloco}. These are descriptive external
references rather than matched rankings.
The all-method LoCoMo comparison is Table~\ref{tab:locomo}; judge details are in
Appendix~\ref{app:statistics}.

\begin{table}[!t]
\centering
\small
\setlength{\tabcolsep}{3.5pt}
\begin{tabular}{@{}lccccc@{}}
\toprule
\textbf{Method}
& \textbf{RULER}$\uparrow$
& \textbf{LongEval}$\uparrow$
& \textbf{LongBench}$\uparrow$
& \textbf{BABILong}$\uparrow$
& \textbf{LoCoMo Judge}$\uparrow$ \\
& 15-cell & 8k--128k & 6-QA F1 & qa1/2/5 macro & full set \\
\midrule
CoMem + LoRA ($j{=}12$)
& 97.05 & 69.0 & 12.15 & 50.43 & 38.27 \\
\midrule
\multicolumn{6}{@{}l}{\emph{Unmatched descriptive references; do not compare columns as a ranking}} \\
KV-Direct
& 78.80 & 65.2 & 12.17 & 63.00 & 34.59 \\
InfLLM
& 77.83 & 21.6 & 11.86 & 59.52 & 22.21 \\
StreamingLLM
& 23.37 & 30.8 & 11.11 & 48.14 & 25.63 \\
MemoryLLM$^{\dagger}$
& 16.55 & 13.6 & 9.01 & 30.00 & 16.11 \\
\bottomrule
\end{tabular}
\caption{\textbf{Cross-benchmark performance of CoMem and external references.}
The RULER column uses one common 15-cell task set. External rows differ from
CoMem in interface, context treatment, adaptation, and sometimes backbone, so
they are descriptive rather than matched comparisons.
$^{\dagger}$MemoryLLM uses a released 7B model.}
\label{tab:overview}
\end{table}

\subsubsection{RULER and length extrapolation}
Table~\ref{tab:h2h} reports the complete common-support RULER grid; the
matched positional-scaling comparison follows in Table~\ref{tab:scaling}.
\begin{table}[!ht]
\centering
\small
\setlength{\tabcolsep}{3.5pt}
\begin{tabular}{@{}llccccc@{}}
\toprule
\textbf{Method} & \textbf{Task} & \textbf{8k} & \textbf{16k} & \textbf{32k} & \textbf{64k} & \textbf{128k} \\
\midrule
\multirow{3}{*}{\textbf{CoMem + LoRA}}
& single-2 & 100 & 100 & 99 & 99 & 100 \\
& multikey & 95 & 94 & 97 & 91 & 93 \\
& var-track & 96.6 & 97.6 & 98.8 & 99.0 & 95.8 \\
\midrule
\multirow{3}{*}{CoMem frozen ($j{=}9$)}
& single & 96 & 99 & 99 & 99 & 96 \\
& multikey & 44 & 44 & 59 & 30 & 48 \\
& var-track & 45 & 38.8 & 36.4 & 31.2 & 25.8 \\
\midrule
\multirow{3}{*}{KV-Direct}
& single & 100 & 100 & 100 & 100 & 0 \\
& multikey & 100 & 100 & 98 & 88 & 0 \\
& var-track & 100 & 100 & 99.8 & 96.2 & 0 \\
\midrule
\multirow{3}{*}{InfLLM}
& single & 100 & 100 & 92 & 61 & 57 \\
& multikey & 100 & 79 & 65 & 45 & 20 \\
& var-track & 100 & 96.8 & 90.6 & 81.8 & 79.2 \\
\midrule
\multirow{3}{*}{StreamingLLM}
& single & 85 & 36 & 20 & 10 & 2 \\
& multikey & 83 & 34 & 18 & 13 & 4 \\
& var-track & 41.8 & 2.2 & 0.6 & 0.2 & 0.8 \\
\midrule
\multicolumn{7}{@{}l}{\emph{External persistent-memory reference (different backbone)}} \\
\multirow{3}{*}{MemoryLLM$^{\dagger}$}
& single & 29 & 40 & 37 & 21 & 21 \\
& multikey & 28 & 24 & 25 & 12 & 10 \\
& var-track & 0 & 1.2 & 0 & 0 & 0 \\
\bottomrule
\end{tabular}
\caption{Full chat-template-free RULER comparison using official
\texttt{string\_match\_all} ($n{=}100$ per cell). KV-Direct places the full
context in the read with native RoPE and no YaRN; its 64k/128k cells are
unextended stress references rather than a length-extended upper bound. CoMem
keeps a bounded retrieved pack. This is \textbf{Cohort A}: every method uses
\texttt{niah\_single\_2}, \texttt{niah\_multikey\_1}, and
\texttt{variable\_tracking}, with $n{=}100$ and the official scorer. Its CoMem
macro is 97.05. Cohort B instead uses \texttt{niah\_single\_3} to match the
YaRN arm (Table~\ref{tab:scaling}) and yields 96.07; the cohorts are never pooled
in a direct comparison. $^{\dagger}$MemoryLLM is the principal external
persistent-memory reference and uses its released Llama-3-8B-chat checkpoint;
its different backbone precludes matched attribution.}
\label{tab:h2h}
\end{table}

The matched positional-scaling grid is Table~\ref{tab:scaling}; the natural-task breakdowns begin in Tables~\ref{tab:longeval}--\ref{tab:locomo}.
\begin{table}[!t]
\centering
\small
\setlength{\tabcolsep}{3.8pt}
\begin{tabular}{@{}llccccc@{}}
\toprule
\textbf{Task} & \textbf{Method} & \textbf{8k} & \textbf{16k} & \textbf{32k} & \textbf{64k} & \textbf{128k} \\
\midrule
\multirow{4}{*}{single needle}
& KV-Direct & 100 & 100 & 100 & 100 & 0 \\
& KV-Direct+YaRN$^\dagger$ & 100 & 100 & 100 & 100 & \textbf{100} \\
& CoMem+LoRA & 100 & 91 & 97 & 98 & 98 \\
& CoMem+LoRA+YaRN$^\dagger$ & 100 & 90 & 97 & 97 & 98 \\
\midrule
\multirow{4}{*}{multikey}
& KV-Direct & 100 & 100 & 98 & 88 & 0 \\
& KV-Direct+YaRN$^\dagger$ & 98 & 94 & 96 & 91 & 89 \\
& CoMem+LoRA & 94 & 91 & 99 & 90 & \textbf{93} \\
& CoMem+LoRA+YaRN$^\dagger$ & 82 & 85 & 94 & 88 & \textbf{93} \\
\midrule
\multirow{4}{*}{variable tracking}
& KV-Direct & 100 & 100 & 99.8 & 96.2 & 0 \\
& KV-Direct+YaRN$^\dagger$ & 99.2 & 99.4 & 26.6 & 67.2 & 57.8 \\
& CoMem+LoRA & 96.2 & 98.0 & 98.2 & 98.6 & \textbf{99.0} \\
& CoMem+LoRA+YaRN$^\dagger$ & 88.6 & 90.0 & 89.4 & 95.0 & 93.6 \\
\bottomrule
\end{tabular}
\caption{Matched-$n{=}100$ Qwen3-8B RULER scaling. $^\dagger$YaRN factor 4
gives an effective 163,840-token window. At 128k, unextended-backbone CoMem+LoRA scores
98/93/99.0 on single-needle, multikey, and variable tracking, versus
100/89/57.8 for KV-Direct+YaRN. Applying YaRN to CoMem changes these to
98/93/93.6, showing an in-window position-scaling tax on variable tracking.
KV-Direct+YaRN is stronger on 128k single-needle and on several 8k--32k
single/multikey cells; CoMem's advantage is concentrated in variable tracking
and bounded read without full-prompt prefill.}
\label{tab:scaling}
\end{table}

\begin{table}[!htbp]
\centering
\small
\setlength{\tabcolsep}{4pt}
\begin{tabular}{@{}lcccccc@{}}
\toprule
\textbf{Method} & \textbf{8k} & \textbf{16k} & \textbf{32k} & \textbf{64k} & \textbf{128k} & \textbf{Mean} \\
\midrule
Raw replay ($j{=}0$) & 100 & 96 & 99 & 94 & 97 & 97.2 \\
CoMem + LoRA & 69 & 75 & 64 & 67 & 70 & 69.0 \\
CoMem frozen ($j{=}9$) & 8 & 0 & 4 & 0 & 4 & 3.2 \\
KV-Direct & 100 & 96 & 92 & 38 & 0 & 65.2 \\
StreamingLLM & 86 & 34 & 18 & 10 & 6 & 30.8 \\
InfLLM & 60 & 30 & 12 & 4 & 2 & 21.6 \\
MemoryLLM$^{\dagger}$ & 22 & 22 & 16 & 6 & 2 & 13.6 \\
\bottomrule
\end{tabular}
\caption{LongEval register-content accuracy. The raw-replay row reports the
measured length-specific $j{=}0$ results ($n{=}100$/length), not the overview
aggregate repeated across cells. Unmarked rows use the unified chat-template-free
protocol. Means use the common 8k--128k support. KV-Direct
uses native RoPE without YaRN, so its 64k/128k cells are unextended stress
references and the five-length mean is not a length-extended upper bound.
CoMem's separate six-length headline is 72.83 after
including its 4k score of 92. The frozen CoMem control used 48 generation
tokens rather than 16; both budgets exceed the single-number answer length.
$^{\dagger}$MemoryLLM is the external persistent-memory reference and uses a
different chat-tuned backbone, so its row is not a matched comparison.}
\label{tab:longeval}
\end{table}

\begin{table}[!htbp]
\centering
\small
\setlength{\tabcolsep}{3pt}
\begin{tabular}{@{}llrrrrrrrr@{}}
\toprule
\textbf{Method} & \textbf{Task}
& \textbf{0k} & \textbf{1k} & \textbf{2k} & \textbf{4k}
& \textbf{8k} & \textbf{16k} & \textbf{32k} & \textbf{Mean} \\
\midrule
\multirow{3}{*}{\textbf{CoMem + LoRA}}
& qa1 & \textbf{98} & 80 & 68 & 68 & 46 & 17 & 12 & 55.6 \\
& qa2 & 26 & 44 & 43 & 44 & 23 & 8 & 1 & 27.0 \\
& qa5 & 68 & \textbf{76} & \textbf{76} & \textbf{75}
      & 68 & \textbf{60} & \textbf{58} & \textbf{68.7} \\
\midrule
\multirow{3}{*}{CoMem frozen ($j{=}9$)}
& qa1 & \textbf{98} & 59 & 71 & 57 & 7 & 1 & 4 & 42.4 \\
& qa2 & 53 & 17 & 28 & 26 & 8 & 4 & 1 & 19.6 \\
& qa5 & 70 & 74 & 65 & 60 & 41 & 39 & 40 & 55.6 \\
\midrule
\multirow{3}{*}{KV-Direct}
& qa1 & \textbf{98} & \textbf{84} & 80 & \textbf{74}
      & \textbf{80} & \textbf{72} & \textbf{63} & \textbf{78.7} \\
& qa2 & \textbf{58} & 53 & \textbf{50} & 46
      & \textbf{49} & \textbf{49} & \textbf{37} & \textbf{48.9} \\
& qa5 & 71 & 73 & 62 & 59 & 65 & 42 & \textbf{58} & 61.4 \\
\midrule
\multirow{3}{*}{InfLLM}
& qa1 & 97 & \textbf{84} & \textbf{81} & 72 & 69 & 52 & 34 & 69.9 \\
& qa2 & 51 & \textbf{58} & 41 & 47 & 41 & 39 & 30 & 43.9 \\
& qa5 & 69 & 68 & 63 & 66 & \textbf{73} & \textbf{60} & 55 & 64.9 \\
\midrule
\multirow{3}{*}{StreamingLLM}
& qa1 & 97 & 83 & 80 & 71 & 23 & 27 & 12 & 56.1 \\
& qa2 & 49 & 56 & 44 & \textbf{49} & 19 & 11 & 4 & 33.1 \\
& qa5 & 68 & 65 & 68 & 65 & 39 & 47 & 34 & 55.1 \\
\midrule
\multirow{3}{*}{MemoryLLM$^{\dagger}$}
& qa1 & 52 & 46 & 35 & 28 & 23 & 17 & 12 & 30.4 \\
& qa2 & 37 & 29 & 17 & 18 & 21 & 17 & 11 & 21.4 \\
& qa5 & 50 & 45 & 39 & 35 & 35 & 34 & 29 & 38.1 \\
\bottomrule
\end{tabular}
\caption{BABILong accuracy ($n{=}100$ per task--length cell), evaluated with
the official \texttt{compare\_answers} scorer and \texttt{TASK\_LABELS}.
Within the native context window, full-context KV-Direct is substantially
stronger on qa1 and qa2, exposing the compression tax, whereas distilled CoMem
achieves the highest qa5 mean, ahead of InfLLM (68.7 vs.\ 64.9).
All methods are evaluated under the unified plain-text prompting protocol,
with model-specific chat templating disabled
(\texttt{chat\_template=False}).
$^{\dagger}$MemoryLLM is the principal external persistent-memory reference and
uses its released Llama-3-8B-chat checkpoint. Its different backbone precludes
matched attribution. We retain the uniform no-chat row and make no
prompt-sensitivity claim without a paper-retained matched table.}
\label{tab:babilong}
\end{table}

\begin{table}[!htbp]
\centering
\small
\setlength{\tabcolsep}{3.5pt}
\begin{tabular}{@{}lrrrrrrr@{}}
\toprule
\textbf{Method}
& \textbf{NQA}
& \textbf{Qasper}
& \textbf{Hotpot}
& \textbf{2Wiki}
& \textbf{MultiF.}
& \textbf{Musique}
& \textbf{Macro} \\
\midrule
Raw replay ($j{=}0$)
& 3.88
& 11.73
& 12.17
& 12.42
& 26.18
& 7.47
& 12.31 \\

CoMem + LoRA
& 4.12
& 11.01
& 11.62
& 12.83
& 25.41
& 7.91
& 12.15 \\

CoMem frozen ($j{=}9$)
& 4.63
& 11.23
& 9.41
& 10.71
& 22.05
& 5.72
& 10.63 \\

KV-Direct
& 3.70
& 11.82
& 12.68
& 12.03
& 25.30
& 7.49
& 12.17 \\

InfLLM
& 2.99
& 11.35
& 12.08
& 12.50
& 25.33
& 6.94
& 11.86 \\

StreamingLLM
& 3.49
& 11.64
& 11.02
& 12.19
& 22.42
& 5.88
& 11.11 \\

MemoryLLM$^{\dagger}$
& 3.13
& 8.46
& 8.76
& 10.27
& 17.43
& 5.98
& 9.01 \\

LLoCO$^{\ddagger}$
& 24.21
& 24.45
& 44.24
& ---
& ---
& ---
& 30.97 \\
\bottomrule
\end{tabular}
\caption{Official LongBench QA token F1 by dataset. The raw-replay row contains
the measured per-dataset $j{=}0$ scores ($n{=}150$--200/dataset), with 12.31 as
their six-task macro. Except for the explicitly marked LLoCO diagnostic, methods
use the same plain-text no-chat protocol. CoMem with LoRA scores
12.15 and full-context KV-Direct scores 12.17; because only CoMem uses the
split-interface adapter, this near-equality is descriptive rather than a
training-budget-matched compression estimate. The frozen
CoMem configuration reports adapter-free performance at $j{=}9$.
NQA denotes NarrativeQA, and MultiF. denotes MultiFieldQA-en.
$^{\dagger}$MemoryLLM is the principal external persistent-memory reference and
uses its released Llama-3-8B-chat checkpoint. Its different backbone precludes
matched attribution. We retain the uniform no-chat row and make no
prompt-sensitivity claim without a paper-retained matched table.
$^{\ddagger}$LLoCO uses LLaMA-2-7B AutoCompressor, released per-domain supervised
LoRAs, fp16, and its native task prompts. Released weights cover only
NarrativeQA/Qasper/HotpotQA on our 200-example sets; 30.97 is their three-task
macro and is not comparable to the six-task macro column.}
\label{tab:longbench}
\end{table}

\begin{table}[!t]
\centering
\small
\setlength{\tabcolsep}{4pt}
\begin{tabular}{@{}lcccc@{}}
\toprule
\textbf{Method} & \textbf{Judge} & \textbf{Judge$_{1:4}$} & \textbf{F1} & \textbf{Substr. acc.} \\
\midrule
Raw replay ($j{=}0$) & 41.59 & 52.60 & 9.90 & 25.23 \\
CoMem + LoRA ($j{=}12$) & 38.27 & 48.64 & 9.15 & 23.36 \\
KV-Direct & 34.59 & 43.83 & 9.02 & 22.36 \\
StreamingLLM & 25.63 & 31.04 & 7.67 & 13.75 \\
InfLLM & 22.21 & 26.43 & 7.39 & 13.34 \\
MemoryLLM$^{\dagger}$ & 16.11 & 20.65 & 5.91 & 9.52 \\
\bottomrule
\end{tabular}
\caption{LoCoMo comparison on all 1,986 chat-template-free examples. The
raw-replay row is the measured full-set $j{=}0$ reference rather than an
overview aggregate duplicated into category cells.
Judge$_{1:4}$ is GPT-4o accuracy on the 1,540 answerable items; the full Judge
column folds in local abstention scoring for 446 adversarial items. F1 and
substring accuracy are deterministic lexical diagnostics; semantic Judge is the
primary metric. All rows use the same scoring protocol. The same-$j$ adapter effect is isolated in
Table~\ref{tab:depth-tradeoff}. On paired answerable items, distilled CoMem
exceeds KV-Direct by $4.81$ points (item-bootstrap 95\% CI $[2.34,7.34]$);
a 10-conversation cluster bootstrap gives $[1.27,8.32]$ and 8/10 observed
conversation differences favor CoMem. An independent DeepSeek-V3 audit
preserves the ordering.
$^{\dagger}$MemoryLLM is the external persistent-memory reference and uses its
official chat-tuned Llama-3 checkpoint; its different backbone precludes matched
attribution. The paper retains only its uniform no-chat evaluation and makes no
native-chat sensitivity claim.}
\label{tab:locomo}
\end{table}

\subsection{Store scaling and failure boundaries}
\label{app:scaling}
Table~\ref{tab:store-scaling} tests the ``bounded read over extensible store''
claim directly. Retrieval is invariant when the required evidence fits top-12,
but lookup and index construction still scale with store size. Beyond the budget,
recall is physically capped; co-keyed multivalue evidence falls below even this
cap (recall 0.62 at $E{=}8$), and a common-word-frequency task scores zero as
coverage falls from 4.5\% at 128k to 1.2\% at 512k. These negative results delimit
the claim: bounded retrieval is not global aggregation, and successful retrieval
does not guarantee multi-hop reasoning.

\begin{table}[t]
\centering
\small
\setlength{\tabcolsep}{3pt}
\begin{tabular}{@{}lrrrr@{}}
\toprule
\textbf{Store} & \textbf{NIAH rec.} & \textbf{NIAH score} &
\textbf{VT rec.} & \textbf{VT score} \\
\midrule
128k & 1.00 & 100 & 1.00 & 18 \\
256k & 1.00 & 100 & 1.00 & 34 \\
512k & 1.00 & 100 & 1.00 & 40 \\
1M   & 1.00 & 100 & 1.00 & 54 \\
2M   & 1.00 & 90  & 1.00 & 38 \\
4M   & 1.00 & 100 & 1.00 & 34 \\
\bottomrule
\end{tabular}

\vspace{3pt}
\begin{tabular}{@{}lrr@{}}
\toprule
\textbf{Store} & \textbf{Read tokens} & \textbf{BM25 (ms)} \\
\midrule
128k & 6,211 / 6,463 & 80 / 20 \\
256k & 6,209 / 6,463 & 164 / 39 \\
512k & 6,209 / 6,464 & 328 / 81 \\
1M   & 6,209 / 6,463 & 697 / 170 \\
2M   & 6,211 / 6,463 & 1,407 / 357 \\
4M   & 6,208 / 6,463 & 2,852 / 725 \\
\bottomrule
\end{tabular}
\caption{Store scaling with a fixed top-12 read. Values separated by ``/'' are
single-evidence NIAH / five-link variable tracking (VT). Evidence recall
($n{=}20$) stays perfect and the model read stays about 6.2--6.5k tokens from
128k to 4M stored tokens; BM25 lookup grows approximately linearly. Generation
scores use $n{=}10$ and reveal that VT reasoning remains noisy even when all
five evidence chunks are retrieved. At 128k, increasing required VT evidence
from 12 to 16/24/32 gives recall 1.0 to 0.75/0.50/0.375, exactly
$\min(1,12/E)$.}
\label{tab:store-scaling}
\end{table}

We audited the PG-19 adapter corpus against the InfiniteBench long-book support
using normalized 13-gram containment sketches. The conservative 0.8 threshold
flags 24/86 books (163/580 records), but containment is sharply bimodal: 31 books
are below 0.10, no book lies in 0.18--0.60, and 54 are at least 0.60. Any cut in
0.20--0.60 therefore flags 54/86 books and 387/580 records; anonymized character
names make the 0.8 count an undercount. Existing predictions were unavailable
for a strict-clean-subset rescore (113 QA and 76 choice records), so we remove
the book-quality comparison rather than treat it as out-of-domain evidence.
This does not affect the synthetic store-scaling results above.
We subsequently audited the PG-19 adapter corpus against the natural-task
suites. LoCoMo has no flagged overlap and LongEval is synthetic by construction.
Within the six LongBench QA datasets, material overlap is localized to
NarrativeQA: 96 of 200 examples are removed by the strict 13-gram clean filter,
leaving 104. Recomputing all available rows changes six-task macro F1 by at most
0.13 points; for example, frozen CoMem changes 10.63$\rightarrow$10.60 and
KV-Direct 10.07$\rightarrow$9.94. The PG-19-trained $j{=}12$ row changes
9.58$\rightarrow$9.57. Thus the displayed LongBench ordering and near-ties are
robust to this strict subset, although the audit detects near-verbatim overlap
rather than semantic paraphrase or exposure through other pretraining corpora.

\subsection{Reproducibility details}
\label{app:reproducibility}
Tables~\ref{tab:repro-config}--\ref{tab:repro-eval} record the configuration,
training, and evaluation settings used by the Qwen3-8B flagship. The archived
configuration files and evaluation scripts accompany the code release.

\begin{table}[!t]
\centering
\small
\renewcommand{\arraystretch}{0.92}
\setlength{\tabcolsep}{3pt}
\begin{tabular}{@{}p{0.14\textwidth}p{0.20\textwidth}p{0.60\textwidth}@{}}
\toprule
\textbf{Layer} & \textbf{Setting} & \textbf{Value} \\
\midrule
Backbone & Architecture & Frozen Qwen/Qwen3-8B; revision \texttt{b968826d9c46} (full hash in artifact manifest); $L{=}36$, $d{=}4096$, 32 query heads, 8 KV heads, bf16, SDPA. \\
Backbone & Positions & Native window 40,960; RoPE $\theta{=}10^6$; no active RoPE scaling or YaRN. \\
Write & Cached state & Split $j{=}12$; disjoint 512-token chunks; chunk-local positions restart at zero; one bf16 residual tensor $h_j$ per token. \\
Read & Pack & $[\mathrm{BOS};\,\text{selected }h_j;\,h_j(q)]$ in document order, with fresh contiguous RoPE and full causal cross-chunk attention. The nominal cap is $1+12\!\times\!512+512=6{,}657$ tokens; actual RULER means are about 6.2--6.5k because tail chunks and queries can be shorter. \\
Retrieval & BM25 & Token-ID documents and bare-question token IDs; $k_1{=}1.5$, $b{=}0.75$, Robertson IDF with $+1$ smoothing; no lowercasing, stemming, or stop-word removal. \\
Retrieval & Iteration & \texttt{rounds=0} means automatic $\lceil k/h\rceil$ rounds, not one-shot. Each round uses newly selected chunks as the next frontier and deduplicates the selected set. \\
Retrieval & Width & All five benchmarks fix $k{=}12$, hop width $h{=}4$, and therefore three automatic rounds. Chunk size 512 and the BOS sink are also fixed. \\
Read & Mask & The pack is one causal sequence: the final query sees all chunks; later chunks may attend to earlier chunks. In the block-diagonal ablation, chunks see only themselves and the sink, while the final query still sees all chunks. \\
Generation & Decoding & Greedy decoding (no sampling, temperature, top-$p$, or beams); no chat template and no thinking mode. \\
Adapter-free arm & Deployment & Frozen backbone, no LoRA, $j{=}9$; all other memory and evaluation settings are held fixed. \\
\bottomrule
\end{tabular}
\caption{Flagship CoMem inference configuration. The retrieval and read
configuration is frozen across all five benchmarks; only benchmark harness
parameters differ.}
\label{tab:repro-config}
\end{table}

\begin{table}[!t]
\centering
\small
\renewcommand{\arraystretch}{0.92}
\setlength{\tabcolsep}{3pt}
\begin{tabular}{@{}p{0.18\textwidth}p{0.76\textwidth}@{}}
\toprule
\textbf{Setting} & \textbf{Value} \\
\midrule
Adapter & Rank 32, $\alpha{=}64$, dropout 0; Q/K/V/O and gate/up/down projections in layers 12--35 only; 58.20M trainable parameters (0.71\%). Final weight SHA-256: \path{dd09cd17457c63578c0f38dab79b287ab5da6e3f14c119aedafec1c34400536f}. \\
Teacher and objective & The same Qwen3-8B with adapters disabled and $j{=}0$ serves as the frozen teacher. At every query token, the teacher top-64 vocabulary indices define a shared support; teacher and gathered student logits are each softmax-renormalized on that support, and outside-support logits are discarded. $\mathcal L=0.6\,\mathrm{KL}(p\Vert q)+0.4\,\mathrm{KL}(q\Vert p)$ at temperature 1, averaged over query tokens, with no CE term. Retained teacher top-64 probability mass was not logged. \\
Data & Plain PG19 train text; no synthetic long-context examples, retrieval, task labels, or benchmark supervision. Training uses 2,048-token windows, formed as four 512-token chunks. \\
Forward construction & Each 2,048-token window is split into three context chunks and one query chunk. Teacher and student use the same BOS sink, chunk token IDs, order, and query segment. The teacher replays every chunk from layer 0 with the adapter disabled; the student writes chunks independently through layers 0--11 without gradient and resumes one causal packed sequence through layers 12--35 with the LoRA. Loss is computed only on all query tokens. \\
Optimization & AdamW, $\beta=(0.9,0.95)$, weight decay 0, peak LR $8\times10^{-5}$, 100-step linear warmup, cosine decay to zero, gradient clipping 1.0, seed 42, bf16, no gradient checkpointing. \\
Schedule & 4,000 steps on 8 GPUs, one sample per GPU and no gradient accumulation (global batch 8), totaling about 65.5M tokens. \\
Compute & One node with 8$\times$NVIDIA H20 GPUs. Logged throughput is about 24.5 samples/s, implying roughly 22 minutes wall-clock; training peak memory was not recorded. \\
\bottomrule
\end{tabular}
\caption{Self-distillation and LoRA training configuration. The adapter name's
``4k'' denotes 4,000 optimization steps; the training sequence length is 2,048
tokens.}
\label{tab:repro-train}
\end{table}

\begin{table}[!t]
\centering
\small
\renewcommand{\arraystretch}{0.92}
\setlength{\tabcolsep}{3pt}
\begin{tabular}{@{}p{0.13\textwidth}p{0.25\textwidth}p{0.15\textwidth}p{0.39\textwidth}@{}}
\toprule
\textbf{Benchmark} & \textbf{Tasks and support} & \textbf{Generation} & \textbf{Scoring and aggregation} \\
\midrule
RULER & Single needle, multikey, and variable tracking; 8k--128k; 100 examples/cell; 8 shards; seed 42. & 48 tokens; 60 for variable tracking. & Official case-insensitive \texttt{string\_match\_all}; equal-weight macro over 15 cells. \\
LongEval & Lines retrieval; 4k--128k; 100 examples/length; 8 shards; seed 1234. & 16 tokens for the flagship; 48 for frozen controls. & First numeric answer exact match; six-length macro (five-length 8k--128k macro for matched baseline tables). \\
LongBench & Six QA datasets; 1,150 examples total; 8 shards; seed 42. & 32/64/128 tokens by dataset. & Official multi-reference token F1; equal-weight macro over six datasets. \\
BABILong & qa1/qa2/qa5; 0k--32k; 100 examples/cell; 4 shards; seed 42. & 20 tokens. & Official \texttt{compare\_answers} with \texttt{TASK\_LABELS}. \\
LoCoMo & All 1,986 QA pairs from 10 conversations; 4 shards. & 48 tokens for the main 8B comparison; 512 for the cross-scale audit. & GPT-4o judge on categories 1--4; official local abstention rule for category 5; full-set judge is the headline. \\
\bottomrule
\end{tabular}
\caption{Benchmark-level evaluation settings. Every main comparison uses
\texttt{chat\_template=False}. MemoryLLM is the external persistent-memory
reference, evaluated with its released chat-tuned backbone.}
\label{tab:repro-eval}
\end{table}

\paragraph{Artifacts, licenses, and data scope.}
The study evaluates English text only. It uses the frozen released
Qwen/Qwen3-8B checkpoint (Apache-2.0); we do no backbone continued pretraining.
It also uses PG-19~\citep{pg19}, RULER~\citep{ruler}, BABILong~\citep{babilong},
LongBench~\citep{longbench}, LongEval/LongChat~\citep{longchat}, and
LoCoMo~\citep{locomo}. RULER and LongChat code are Apache-2.0; the LongBench
repository is MIT-licensed, while its component datasets retain their upstream
terms; BABILong code is Apache-2.0 and incorporates BSD-licensed bAbI tasks;
LoCoMo is CC BY-NC 4.0. PG-19 contains pre-1919 Project Gutenberg books, for
which the dataset page does not assert one uniform corpus license, so we do not
redistribute its text. We likewise redistribute no model weights, benchmark
examples, or private API responses. Subject to upstream terms, the anonymous
archive includes the adapter, source, documentation, pinned requirements,
prediction hashes or permissible shards, judge prompts/parsed outputs, raw
timing records, and third-party notices. Users obtain all external artifacts
from their original providers and remain
responsible for their terms. Table~\ref{tab:repro-eval} reports the evaluated
supports and sample counts; LoCoMo category denominators are in
Appendix~\ref{app:statistics}.

\paragraph{Software environment.}
The archived environment specifies Python 3.10+, PyTorch 2.10,
Transformers 5.5.4, PEFT 0.10+, Datasets 2.14+, bf16, and SDPA. Evaluation uses
the benchmark-specific official scorers named in Table~\ref{tab:repro-eval}; the
archive records all non-default flags and provides a CPU correctness gate.

\paragraph{Compute accounting.}
The final adapter run used one node with 8 H20 GPUs for about 22 minutes, or
approximately 2.9 H20 GPU-hours. Table~\ref{tab:repro-train} and this section
report model size, trainable parameters, hardware, and measured timing protocol.
Total GPU-hours across preliminary probes, failed runs, baseline generation, and
all ablations were not consistently logged; we report this omission rather than
extrapolating an unreliable total.

\subsection{Efficiency, replay attribution, and robustness}
\paragraph{Efficiency measurement.}
\label{app:efficiency}
The Write-inclusive cohort uses one NVIDIA L20A (183\,GB HBM) with PyTorch 2.10,
Transformers 5.5.4, bf16, SDPA, and the rank-32 LoRA enabled, reporting medians
of three repetitions. Full-context prefill times the complete forward; CoMem
times document Write, CPU retrieval, query Write, and upper-layer Read through
logits. The timer excludes one-time BM25 index construction and external-store
I/O; Table~\ref{tab:store-io} measures the latter separately. At 128k,
full-context prefill is 6.035\,s and peaks at 89.39\,GB, while CoMem takes
2.202\,s and peaks at 18.54\,GB. Cached decode uses 20 greedy steps.

\begin{table}[!t]
\centering
\small
\setlength{\tabcolsep}{3.3pt}
\begin{tabular}{@{}lccccc@{}}
\toprule
\textbf{Length} & \shortstack{\textbf{Full}\\\textbf{time}} & \shortstack{\textbf{CoMem}\\\textbf{time}} & \textbf{Speedup} & \shortstack{\textbf{Full}\\\textbf{mem.}} & \shortstack{\textbf{CoMem}\\\textbf{mem.}} \\
\midrule
8k   & 0.165\,s & 0.309\,s & 0.53$\times$ & 19.92\,GB & 17.60\,GB \\
16k  & 0.356\,s & 0.441\,s & 0.81$\times$ & 24.55\,GB & 17.67\,GB \\
32k  & 0.826\,s & 0.681\,s & 1.21$\times$ & 33.82\,GB & 17.79\,GB \\
64k  & 2.104\,s & 1.200\,s & 1.75$\times$ & 52.34\,GB & 18.04\,GB \\
128k & 6.035\,s & 2.202\,s & \textbf{2.74$\times$} & 89.39\,GB & \textbf{18.54\,GB} \\
\bottomrule
\end{tabular}
\caption{Same-platform Qwen3-8B pipeline efficiency with the same distilled rank-32 LoRA
attached to both paths (one NVIDIA L20A, 183\,GB HBM, bf16/SDPA, median of three). This
is a separate cohort from the stock-dense H20 measurement in
Table~\ref{tab:online-prefill}. ``Full''
processes the entire context; ``CoMem'' includes document Write, CPU retrieval,
query Write, and upper-layer Read over the selected pack, but excludes one-time
BM25 index construction and external-store I/O. The select-first online path is
reported separately in Table~\ref{tab:online-prefill}; it assumes a prewritten
store and is not mixed with these full-pipeline measurements.}
\label{tab:eff}
\end{table}

The store-ready online-prefill cohort above is the paper's only dense-context
online reference. The dense arm is stock Qwen3-8B without the adapter, while
CoMem uses the flagship LoRA, so it is a composed deployment point rather than
an adapter-matched depth estimate. The same-adapter L20A pipeline in
Table~\ref{tab:eff} is the complementary Write-inclusive reference.

The main serving experiment instead measures write-once/fetch/reuse directly.
At 32k, CPU-pinned break-even is 8.9--10.9 queries for generations up to
128 tokens, while the complete GPU/CPU grid is 5.5--10.9. At 128k, the
$G\le128$ crossover spans 25.8--37.4 queries; at 1M it spans about
180--421. The difference from the earlier component
cohort comes from source length, generation length, storage tier, and timing
boundary.
CoMem stores 8\,KiB/token (about
1\,GiB at 128k), far more than text but 18$\times$ less than full KV. Main timing
cohorts keep the store in HBM; Table~\ref{tab:store-io} separately measures
placement beyond HBM, so its I/O microbenchmark is not added to model latency.

\paragraph{Retained-KV references.}
Table~\ref{tab:kvcompress} reports SnapKV and PyramidKV
~\citep{snapkv,pyramidkv} under their full-prefill boundary. They answer a
different systems question from cross-query persistence and are not treated as
nearest modular-cache baselines.

\begin{table}[!t]
\centering
\small
\setlength{\tabcolsep}{3.2pt}
\begin{minipage}[t]{\textwidth}
\centering
\textit{(a) Quality at a 6,657-token/KV budget}\\[-2pt]
\setlength{\tabcolsep}{5pt}
\begin{tabular}{@{}lccccr@{}}
\toprule
\textbf{Method} & \textbf{Native A} & \multicolumn{3}{c}{\textbf{Full-prompt 64k/128k}} & \textbf{LoCoMo} \\
& \textbf{macro} & \textbf{single} & \textbf{multikey} & \textbf{tracking} & \textbf{F1/acc} \\
\midrule
CoMem+LoRA & \textbf{97.05} & 99/100 & 91/93 & 99.0/95.8 & 9.15/23.36 \\
SnapKV & 72.33$^{\dagger}$ & 100/100 & 94/91 & 80.0/84.8 & 9.21/22.05 \\
PyramidKV & 72.32$^{\dagger}$ & 100/100 & 93/89 & 88.6/86.8 & 9.13/22.10 \\
\bottomrule
\end{tabular}
\end{minipage}

\vspace{6pt}
\begin{minipage}[t]{\textwidth}
\centering
\textit{(b) Full-prefill systems cost on one H20}\\[-2pt]
\setlength{\tabcolsep}{6pt}
\begin{tabular}{@{}lrrrr@{}}
\toprule
\textbf{Method} & \textbf{Len.} & \textbf{ms} & \textbf{GB} & \textbf{ms/tok} \\
\midrule
SnapKV & 8k & 1,213 & 18.6 & 24.7 \\
& 32k & 6,342 & 25.7 & 24.6 \\
& 128k & 51,520 & 54.3 & 24.7 \\
PyramidKV & 8k & 1,203 & 18.6 & 24.7 \\
& 32k & 6,301 & 25.7 & 26.7 \\
& 128k & 51,520 & 54.3 & 26.8 \\
\bottomrule
\end{tabular}
\end{minipage}
\caption{Selected/compressed-KV baselines at a similar retained-KV size but a
different full-prefill boundary.
Panel (a) uses RULER Cohort A and full-set LoCoMo lexical scoring. CoMem reads
its bounded pack from the full store without YaRN; the SnapKV/PyramidKV 64k/128k
columns use YaRN factor 4 and see the full prompt. $^{\dagger}$Their native
15-cell macros include 64k/128k inputs left-truncated to the 40,960-token native
window; within 8k--32k both score 99.96. Thus 72.33/72.32 are not
common-native-support quality summaries. Panel (b) times full-prompt prefill,
compression, and 64-token decode (median of three); the final column is decode
latency. These methods must prefill the full prompt before eviction. CoMem's
separate L20A cohort peaks at 18.54\,GB at 128k
(Appendix~\ref{app:efficiency}); hardware differs, so we report no cross-table latency
ratio.}
\label{tab:kvcompress}
\end{table}

\paragraph{Same-backbone full-depth chunk-KV baseline.}
Table~\ref{tab:cacheblend-baseline} uses a minimal faithful CacheBlend-style
implementation on Qwen3-8B: full-depth per-chunk KV, exact global RoPE
reindexing, and selective context-token recomputation. The fp32 real-model
self-test gives maximum KV reindex deviations below
$4.4\!\times\!10^{-5}$ and makes $r{=}1$ agree with full prefill to
$2.8\!\times\!10^{-5}$ maximum logit error with 100\% top-1 agreement. The
released aggregate contains all 48 RULER ratio/task/length cells, 24 BABILong
cells, and four full-set LoCoMo ratio points with no missing required cells.
Full-depth Qwen3 GQA KV is
$2\cdot36\cdot8\cdot128\cdot2=147{,}456$ bytes/token, versus 8,192 bytes/token
for one bf16 residual. The baseline is training-free and does not include
CacheBlend's complete native scheduler or cache manager.

\paragraph{Paired quality--latency replay control.}
Table~\ref{tab:replay-latency} uses a separate H20 harness. The two arms share
the Qwen3-8B checkpoint, flagship LoRA SHA-256 and all 168 mounted modules,
example IDs, decode parameters, and per-example retrieved chunk IDs, order, and
pack tokens; only replay start differs. Across 15 RULER Cohort-B cells
($n{=}100$ each), full replay scores 99.19 and layer-12 replay scores 96.07.
The paired 3.12-point gap has bootstrap 95\% CI $[2.36,3.93]$ and exact McNemar
$p{=}8.79\!\times\!10^{-24}$ (83 full-only correct versus one layer-12-only
correct). The gap is largest on distractor-heavy multikey retrieval.

For latency, each arm uses the same approximately 6.5k-token pack retrieved from
16k source contexts in three independent
processes, with three warmups and 20 reads per process. Layer-12 replay is
$1.403\times$ faster (931.9 to 664.4\,ms), with all three process-level ratios
between 1.402 and 1.404. Total-decode medians remain similar at roughly
2.76--2.86\,s; adding the observed Read and decode components gives only about
$1.07$--$1.09\times$ total ratios. A preceding component harness attributes the read reduction
primarily to upper-transformer forward ($1.398\times$), with unchanged LM-head
time. Retrieval, index construction, reusable store/query Write, and persistent
I/O are outside this read-phase timing. The result is therefore a measured
quality--latency trade-off, not quality-preserving or end-to-end acceleration.

\begin{table}[!htbp]
\centering
\small
\setlength{\tabcolsep}{4pt}
\begin{tabular}{@{}lrrrr@{}}
\toprule
\textbf{Replay start} & \textbf{RULER} & \textbf{Read (ms)} & \textbf{p10} & \textbf{p90} \\
\midrule
$j{=}0$  & \textbf{99.19} & 931.9 & 931.6 & 942.0 \\
$j{=}12$ & 96.07 & \textbf{664.4} & 663.8 & 667.1 \\
\midrule
Trade-off & $-3.12$ pt & \multicolumn{3}{c}{$1.403\times$ speedup} \\
\bottomrule
\end{tabular}
\caption{Paired quality--latency control with the same LoRA. Quality is the
15-cell RULER macro ($n{=}1{,}500$ paired examples); the 3.12-point loss has
paired-bootstrap 95\% CI $[2.36,3.93]$. Latency uses a fixed approximately
6.5k-token pack retrieved from 16k source contexts and reports
medians across three independent processes (20 reads each). Retrieval, reusable
Write, and I/O are excluded. Total decode is similar across arms (about
2.76--2.86\,s), reducing the combined Read-plus-decode ratio to roughly
$1.07$--$1.09\times$.}
\label{tab:replay-latency}
\end{table}

\begin{table}[!htbp]
\centering
\small
\setlength{\tabcolsep}{5pt}
\begin{tabular}{@{}lrrc@{}}
\toprule
\textbf{Replay pathway} & \textbf{RULER B} & \textbf{MK 8/16k} & \textbf{Reusable} \\
\midrule
Full replay from $j{=}0$ & 99.19 & 100.0 & No \\
Continuous-prefix $h_{12}$ oracle & 99.19 & 100.0 & No \\
Chunk-local cached $h_{12}$ & 96.07 & 92.5 & Yes \\
\bottomrule
\end{tabular}
\caption{Write-side attribution with identical examples, packs, and LoRA.
The continuous-prefix oracle runs layers 0--11 over the selected packed sequence
for every query, then starts upper replay from that $h_{12}$; it is bit-identical
to full replay on all 1,500 Cohort-B examples. The deployable chunk-local path
loses 3.12 points (95\% CI $[2.36,3.93]$; McNemar
$p{=}8.79\!\times\!10^{-24}$), and 7.50 points on the pooled 8k/16k multikey
cells ($n{=}200$, CI $[4.0,11.5]$). The oracle is an attribution upper bound,
not a reusable cache.}
\label{tab:h12-oracle}
\end{table}

\begin{table}[!htbp]
\centering
\small
\setlength{\tabcolsep}{3pt}
\resizebox{\columnwidth}{!}{%
\begin{tabular}{@{}lrrrrl@{}}
\toprule
\textbf{Backend} & \textbf{Write GB/s} & \textbf{Fetch ms} & \textbf{H2D ms} & \textbf{Peak QPS} & \textbf{16M outcome} \\
\midrule
GPU resident & OOM & -- & -- & -- & 128 GiB $>$ 95 GiB \\
CPU pinned & 139.6 & 0.81 & 1.41 & 941 & 235 GB host \\
Local NVMe & 2.45 & 13.42 & 0.91 & 264 & 2.18 GB host \\
Network/CEPH & 1.66 & 79.20 & 1.20 & 42 & 2.19 GB host \\
\bottomrule
\end{tabular}%
}
\caption{Measured persistent-store tiers at 16M tokens (128\,GiB of bf16
$h_{12}$). Each query fetches the same 50.3\,MB top-12 pack; medians use seven
repetitions after two warmups, and QPS is the best of 1/4/16 concurrent clients.
GPU residence fits 8M tokens (64.98\,GB peak) but not 16M; off-GPU tiers retain
store-size-independent transfer volume at increasing I/O latency. This
microbenchmark times storage and transfer, not BM25 or model inference.}
\label{tab:store-io}
\end{table}

\subsection{Seed-plus-effective-batch robustness}

Table~\ref{tab:seed-batch-robustness} reports the available run-level
aggregates. Across the 18 individual RULER and BABILong cells, the median
cell-wise SD is 1.34 points and the maximum is 4.36. Because the two added runs
change effective batch from 8 to 3, these numbers jointly probe initialization
and optimization-noise robustness; they are not a clean three-seed variance
estimate. The available evidence also does not support a multi-run claim for the
exact 15-cell RULER-B, LoCoMo, or LongEval headlines.

\begin{table}[!t]
\centering
\small
\setlength{\tabcolsep}{4pt}
\begin{tabular}{@{}llrrrrr@{}}
\toprule
\textbf{Suite} & \textbf{Adapter} & \textbf{Seed 42} & \textbf{Seed 1} &
\textbf{Seed 2} & \textbf{Mean} & \textbf{SD} \\
\midrule
RULER 9-cell macro & batch 8 / 3 / 3 & 98.80 & 99.42 & 98.76 & 98.99 & 0.37 \\
BABILong 9-cell macro & batch 8 / 3 / 3 & 38.44 & 40.33 & 41.44 & 40.07 & 1.52 \\
\midrule
RULER multikey 3-length mean & batch 8 / 3 / 3 & 96.67 & 98.67 & 96.67 & 97.33 & 1.15 \\
BABILong qa5 3-length mean & batch 8 / 3 / 3 & 66.00 & 69.33 & 71.00 & 68.78 & 2.55 \\
\bottomrule
\end{tabular}
\caption{\textbf{Seed-plus-effective-batch robustness, not clean seed
variance.} All adapters use the same Qwen3-8B split, PG-19 data budget, and
rank-32 objective; seed 42 is the flagship effective-batch-8 run, while seeds
1 and 2 use effective batch 3 with matched examples seen. Evaluation uses
identical cells and supports: RULER single/multikey/variable-tracking at
32k/64k/128k ($n{=}50$/cell) and BABILong qa1/qa2/qa5 at
4k/16k/32k ($n{=}100$/cell). SD is the sample SD across the three run-level
cell macros. These reduced-support cells are not the 15-cell RULER-B headline,
and no LoCoMo or LongEval multi-run aggregate was retained.}
\label{tab:seed-batch-robustness}
\end{table}

\subsection{Prompt and native-chat sensitivity}

For Qwen3-8B, the GPT-4o LoCoMo judge is substantially less prompt-sensitive
than token F1. CoMem scores 38.27 without chat templating and 37.76 with it;
KV-Direct scores 34.59 and 38.22. In contrast, CoMem's F1/substring accuracy
changes from 9.15/23.36 to 19.51/28.65, while KV-Direct F1 changes from 9.02 to
40.06, showing that lexical metrics are strongly formatting-sensitive. We therefore retain the uniform
no-chat judge comparison as the controlled headline.

For MemoryLLM, we report only the uniform no-chat row in the paper. A matched
native-chat/no-chat table was not retained in the frozen manuscript; therefore
we make no prompt-sensitivity claim for this external baseline and do not mix
prompt protocols within an average.

\subsection{Cross-scale and MoE generality}
\label{app:generality}
Adapter-free sweeps across six Qwen3 sizes and Qwen3-30B-A3B confirm that the
split-forward implementation ports beyond the flagship. Performance is
non-monotonic because split, architecture, and task differ, so we infer neither
a scaling law nor a replicated quality frontier. These exploratory values and
raw predictions are retained in the artifact; the Hy3 study below gives the
stronger architecture-level partition check reported in this appendix.

\subsubsection{Hunyuan Hy3: an 80-layer sparse MoE}
\label{sec:hy3-general}
We additionally port CoMem to Tencent Hunyuan Hy3~\citep{hunyuan}, an 80-layer sparse
Mixture-of-Experts model (192 experts, top-8 routing plus one shared expert;
$\approx597$\,GB in bf16) with a 262,144-token native window. The backbone is
sharded across eight GPUs; Write and Read semantics are unchanged.

\paragraph{Exact partition through MoE routing.}
Reconstructing a contiguous forward by writing \texttt{layers[0:$j$]} and
resuming \texttt{layers[$j$:80]} gives
$\max|\Delta\mathrm{logit}|=0.0$ at $j\in\{0,1,40,80\}$, including the
Dense-to-MoE boundary. The resumed half re-executes every router and expert, so
the partition preserves routing exactly on this self-test.

\paragraph{Split depth and self-distillation.}
A PG19 sweep places the best readout region near $j\approx32$--36. At $j=32$,
self-distillation reduces the 16k multiplicative LM tax from 1.496 to 1.078 and
raises top-1 agreement from 0.773 to 0.871
(Table~\ref{tab:hy3distill}).

\begin{table}[!htbp]
\centering
\small
\begin{tabular}{@{}lcccc@{}}
\toprule
 & \multicolumn{2}{c}{ctx 8k} & \multicolumn{2}{c}{ctx 16k} \\
\cmidrule(lr){2-3}\cmidrule(lr){4-5}
\textbf{$j{=}32$} & gap & top1 & gap & top1 \\
\midrule
zero-shot  & 1.378 & 0.823 & 1.496 & 0.773 \\
distilled  & \textbf{1.071} & \textbf{0.895} & \textbf{1.078} & \textbf{0.871} \\
\bottomrule
\end{tabular}
\caption{CoMem self-distillation on Hy3 at split $j{=}32$ (LoRA $r{=}32,
\alpha{=}64$ on \texttt{layers[32:]}, teacher $j{=}0$ RAG recompute, 4000 steps of
plain PG19 KL, no synthetic data). \textbf{gap} is the multiplicative LM tax
(CoMem-readout perplexity / full-context perplexity; $1.0$ is exact); \textbf{top1}
is argmax agreement with the full forward, over matched 16-doc PG19 windows. The
16k LM tax falls from $+49.6\%$ to $+7.8\%$ and top1 rises $0.773\to0.871$,
reproducing the 8B-backbone self-distillation effect on the 80-layer MoE.}
\label{tab:hy3distill}
\end{table}

\paragraph{Bounded read through 256k.}
With the distilled adapter and BM25 top-8, the read stays at
$\approx4.3$--4.6k tokens from 16k through 256k. CoMem reaches 100/98
single/multikey recall at the 256k native ceiling with no OOM
(Table~\ref{tab:hy3ruler}).

\begin{table}[!htbp]
\centering
\small
\begin{tabular}{@{}lccccc@{}}
\toprule
\textbf{Task} & 16k & 32k & 64k & 128k & 256k \\
\midrule
niah\_single\_2   &  98 & 92 & 100 &  98 & \textbf{100} \\
niah\_multikey\_1 & 100 & 94 & 100 & 100 & \textbf{98} \\
\midrule
read length (tok) & 4.3k & 4.6k & 4.5k & 4.4k & 4.5k \\
context (tok)     & 16k  & 32k  & 65k  & 131k & 261k \\
\bottomrule
\end{tabular}
\caption{CoMem on the Tencent Hunyuan \textbf{Hy3} backbone (\texttt{hy\_v3},
80-layer sparse MoE), RULER needle recall ($n{=}50$, official
\texttt{string\_match\_all}), split $j{=}32$, BM25 top-$8$, self-distilled LoRA.
The read length fed to the model stays $\approx$4.3--4.6k tokens while the
context grows $16\times$ (16k$\to$256k); CoMem retains 100/98 recall at the
backbone's 256k native ceiling, with $0$ out-of-memory failures. The context row
reports the mean packed length of the RULER samples.}
\label{tab:hy3ruler}
\end{table}

\section{Chat-Template-Free Statistical Verification}
\label{app:statistics}

All statistics in this appendix were recomputed from the saved prediction
shards on the shared result filesystem, without GPU reruns. The Qwen3-8B
flagship uses the self-distilled $j{=}12$ adapter, chunk 512, top-12 retrieval,
a BOS sink, and no chat template. The released adapter artifact is identified by the SHA-256 and configuration in
Table~\ref{tab:repro-train}. The main RULER directory uses the
\texttt{iter\_bm25} implementation; the saved configuration records
\texttt{rounds=0} and an observed read of approximately 6.5k tokens. Every reported RULER cell contains eight of eight shards, 100
examples, no empty prediction, and zero mismatch between the stored score and a
fresh official-kernel recomputation. BABILong likewise contains 100 valid
examples in every cell.

\subsection{RULER cohort definitions}
\begin{center}
\captionsetup{hypcap=false}
\centering
\small
\setlength{\tabcolsep}{4pt}
\begin{tabular}{@{}lrrrrr@{}}
\toprule
\textbf{Task} & \textbf{8k} & \textbf{16k} & \textbf{32k} & \textbf{64k} & \textbf{128k} \\
\midrule
niah\_single\_3 & 100.0 & 91.0 & 97.0 & 98.0 & 98.0 \\
niah\_multikey\_1 & 94.0 & 91.0 & 99.0 & 90.0 & 93.0 \\
variable\_tracking & 96.2 & 98.0 & 98.2 & 98.6 & 99.0 \\
\bottomrule
\end{tabular}
\captionof{table}{RULER \textbf{Cohort B}, used for the matched length-extension and
$j{=}0$ comparisons. The $n{=}100$ macro is $1441.0/15=96.0667$; the exact
single-needle variant is \texttt{niah\_single\_3}.}
\label{tab:ruler-statistics}
\end{center}

Cohort A is the all-method table (Appendix Table~\ref{tab:h2h}): it substitutes
\texttt{niah\_single\_2}, keeps the same multikey/tracking tasks and $n{=}100$,
and gives CoMem 97.05. Cohort B above was freshly sampled with
\texttt{niah\_single\_3} because that is the exact task available for both YaRN
arms and $j{=}0$; it gives 96.07. Tables mark every RULER aggregate with $A$ or
$B$, never subtract across them, and exclude legacy 256k cells. Table~\ref{tab:scaling}
contains only Cohort B.

\subsection{LoCoMo judge protocol and denominator}
Prediction shards for all methods and the adapter-free CoMem control deduplicate
to all 1,986 LoCoMo items.
GPT-4o judges the 1,540 answerable items in categories 1--4. Category 5 contains
446 adversarial abstention items and is scored locally: an answer is correct iff
it is empty or matches the refusal rule. Thus the canonical full-set judge is
over all 1,986 items, not only the API-judged subset.

\begin{center}
\small
\captionsetup{hypcap=false}
\setlength{\tabcolsep}{3.5pt}
\begin{tabular}{@{}lrrrrrr@{}}
\toprule
\textbf{Method} & \textbf{Cat. 1} & \textbf{Cat. 2} & \textbf{Cat. 3} & \textbf{Cat. 4} & \textbf{Cat. 5} & \textbf{Overall} \\
\midrule
\textbf{CoMem + LoRA} & \textbf{26.95} & \textbf{19.00} & \textbf{30.21} & \textbf{69.32} & 2.47 & \textbf{38.27} \\
CoMem frozen ($j{=}9$) & 18.79 & 13.40 & 27.08 & 53.75 & 1.12 & 29.15 \\
KV-Direct & 24.11 & 18.69 & 25.00 & 62.19 & 2.69 & 34.59 \\
StreamingLLM & 22.70 & 11.21 & 23.96 & 42.21 & 6.95 & 25.63 \\
InfLLM & 18.44 & 14.33 & 25.00 & 33.89 & \textbf{7.62} & 22.21 \\
MemoryLLM & 15.60 & 8.72 & 15.63 & 27.47 & 0.45 & 16.11 \\
\bottomrule
\end{tabular}
\captionof{table}{LoCoMo GPT-4o judge by category. Categories 1--4 contain
282/321/96/841 judged items; Category 5 contains 446 locally scored abstention
items. Overall aggregates all 1,986 examples.}
\label{tab:locomo-categories}
\end{center}

The judge was evaluated on July 22, 2026 using an OpenAI-compatible
chat-completions endpoint with model name \texttt{gpt-4o}, \texttt{seed=1}, and
no client-set temperature or top-$p$. The endpoint does not expose a dated
model snapshot. Saved item-level records contain the item ID, binary parsed
decision, raw response text, and model string; the anonymous artifact includes
the permissible parsed decisions and their hashes rather than API credentials.
Requests are retried four
times with exponential backoff; unparsable responses and API failures are
conservatively scored wrong. The prompt presents the question, gold answer, and prediction; it accepts any
semantic match (including equivalent date phrasing), rejects contradictions,
omissions, empty answers, and refusals, and requires exactly
\texttt{CORRECT} or \texttt{WRONG}. The verbatim template is released with the
evaluation script.
Gold alternatives are joined with ``OR.'' A reply must begin with
\texttt{CORRECT} or \texttt{WRONG}; a fallback substring vote is used before
defaulting to wrong.

On the common GPT-4o-judged items ($n{=}1540$), CoMem scores 48.64 and KV-Direct
43.83. A paired per-item bootstrap of their difference (10,000 resamples,
seed 1234) gives $+4.81$ points with 95\% CI $[2.34,7.34]$. Because the questions
are nested within only 10 conversations, this interval is not our sole
inferential check. A paired conversation-cluster bootstrap, resampling the 10
conversations rather than individual questions (same 10,000 resamples,
seed 1234), gives the same $+4.81$ point estimate with 95\% CI $[1.27,8.32]$ ---
wider than the per-item interval, as expected with only ten clusters; 8 of the
10 observed conversation-level differences favor CoMem. We do not interpret
the bootstrap tail fraction as a calibrated hypothesis-test $p$ value. We report the latter as a
dependence-aware robustness check rather than claiming that 1,540 questions are
independent. The canonical full-set point
scores are 38.27 and 34.59, respectively.

\subsection{Independent-judge audit}
On July 26, 2026, we reran the binary prompt with DeepSeek-V3 on a stratified 200-item subset
shared by CoMem and KV-Direct. Agreement with GPT-4o is 0.81
($\kappa=0.626$). Both judges preserve the ordering: GPT-4o scores
36.5/32.5 ($+4.0$) and DeepSeek-V3 56.0/49.0 ($+7.0$) for CoMem/KV-Direct.
Absolute calibration differs, but the positive gap is judge-robust.

\subsection{Uncertainty and aggregation checks}
\paragraph{Equal-latency dependence reanalysis.}
We reanalyzed the saved anchor scores without model execution. The estimand is
the equal-weight mean of the nine cell mean paired differences
CoMem($k{=}12$) minus replay($k{=}10$). A \emph{stratified fixed-cell}
bootstrap keeps all nine cells and resamples exactly 100 paired examples within
each cell, giving 95\% intervals $[-14.33,-8.78]$ for BM25 and
$[-4.56,2.56]$ for BGE. A \emph{hierarchical} bootstrap first resamples nine
cell labels with replacement, then independently resamples 100 paired examples
inside each selected occurrence; its intervals are $[-18.67,-5.11]$ and
$[-10.67,8.33]$. Dropping one cell at a time gives ranges
$[-13.13,-9.50]$ and $[-3.50,1.75]$, respectively. The original pooled-IID
intervals, $[-14.44,-8.67]$ and $[-4.67,2.67]$, are retained only as
sensitivity. All use 100,000 replicates and deterministic base seed 20260804.
The first 100 LoCoMo examples are all conversation 0, leaving one observed
cluster; a conversation bootstrap is therefore not identifiable. The release
includes the score-only exports (without benchmark text, answers, or
predictions), script, result JSON, and SHA-256 manifests.

Per-method, 1,000-resample item-bootstrap intervals (seed 1234) are
$[36.20,40.58]$ for CoMem's full-set LoCoMo judge, $[32.48,36.71]$ for
KV-Direct's, $[46.04,50.98]$ and $[41.23,46.30]$ for their category-1--4 judge
scores, and $[8.60,9.72]$ and $[8.51,9.61]$ for their F1 scores. These unpaired,
item-level intervals are descriptive; the paired cluster analysis above is the
appropriate dependence-aware comparison.

We do not reuse cell-bootstrap intervals from the superseded RULER cohort or
legacy BABILong predictions. The separate six-length LongEval aggregate is
72.83 over 600 examples (95\% CI $[69.33,76.33]$); the benchmark summary
uses the predefined five-length 8k--128k mean of 69.0.

\end{document}